\pdfoutput=1

\documentclass[11pt]{article}

\usepackage[preprint]{acl}

\usepackage{times}
\usepackage{latexsym}

\usepackage[T1]{fontenc}

\usepackage[utf8]{inputenc}

\usepackage{inconsolata}

\usepackage{microtype}
\usepackage{hyperref}
\usepackage{url}
\usepackage{booktabs}
\definecolor{darkblue}{rgb}{0, 0, 0.5}
\hypersetup{colorlinks=true, citecolor=darkblue, linkcolor=darkblue, urlcolor=darkblue}
\usepackage{multirow}
\usepackage{multicol}
\usepackage{tablefootnote}

\usepackage{amsmath}
\usepackage{amsfonts}
\usepackage{caption}
\usepackage{subcaption}
\usepackage{wrapfig}
\usepackage{xcolor} 
\usepackage{graphicx}
\usepackage{float}

\usepackage{wrapfig}

\usepackage{algorithmic}
\usepackage[linesnumbered,ruled]{algorithm2e}

\title{
Monte Carlo Sampling for Analyzing In-Context Examples
}

\author{
Stephanie Schoch$^{ }$\and Yangfeng Ji$^{ }$\\
$^{ }$ Department of Computer Science\\University of Virginia\\ Charlottesville, VA 22904\\
\texttt{\{sns2gr,yangfeng\}@virginia.edu}\\
}

\begin{document}
\maketitle
\begin{abstract}
Prior works have shown that in-context learning is brittle to presentation factors such as the order, number, and choice of selected examples. However, ablation-based guidance on selecting the number of examples may ignore the interplay between different presentation factors. 
In this work we develop a Monte Carlo sampling-based method to study the impact of number of examples while explicitly accounting for effects from order and selected examples.
We find that previous guidance on how  many in-context examples to select does not always generalize across different sets of selected examples and orderings, and whether one-shot settings outperform zero-shot settings is highly dependent on the selected example. 
Additionally, inspired by data valuation, we apply our sampling method to in-context example selection to select examples that perform well across different orderings. 
We find a negative result, that while performance is robust to ordering and number of examples, there is an unexpected performance degradation compared to random sampling.
\end{abstract}

\section{Introduction}\label{intro}
In-context learning is an emergent ability of LLMs \citep{wei2022emergent, brown2020language} where an LLM learns to perform an unseen task by seeing a number of demonstrations in the context window \citep{NEURIPS2020_1457c0d6}.
While in-context learning has shown significant potential as a way to extract relevant information from an LLM and align the model with user expectations, it has also exhibited brittleness to an assortment of factors. For example, model performance when learning in-context is sensitive to which examples are selected \citep{rubin-etal-2022-learning, liu-etal-2022-makes, wu-etal-2023-self, pmlr-v202-ye23c} as well as their orderings \citep{lu-etal-2022-fantastically, chen-etal-2023-relation, liu-etal-2022-makes, chang-jia-2023-data, guo-etal-2024-makes, wu-etal-2023-self}. 

Another important parameter, the number of examples, has received comparably little attention. Prior works have suggested that one-shot settings outperform zero-shot settings even when a random label is used \citep{min-etal-2022-rethinking}.  Additional ablations have guided this parameter by citing performance plateaus at set numbers of examples \citep{wang-etal-2024-learning, min-etal-2022-rethinking, wu-etal-2023-self}. 
However, it is unclear whether this guidance holds when accounting for other sensitive factors such as different orderings and selected examples.

Previous work on data valuation has shown the efficacy of applying Monte Carlo sampling to evaluate datum contributions under different permutations in fine-tuning settings \citep{pmlr-v97-ghorbani19c, schoch2023data}. Inspired by this, we develop a Monte Carlo sampling-based method to investigate the impact of number of examples while using permutations to account for order and selection of in-context examples. 

Specifically, we utilize Monte Carlo sampling and analyze performance with the addition of each exemplar. We find that performance plateaus at previously suggested numbers of examples do not consistently generalize under different permutations. 
Further, we find that one-shot performance may be more sensitive to the selected example than previously recognized \citep{min-etal-2022-rethinking, brown2020language} and the guidance of one-shot outperforming zero-shot is dependent on the selected example. Finally, we find that using Monte Carlo sampling to select in-context examples may increase robustness to effects from ordering and selected examples, but unexpectedly, does not lead to performance improvements over random sampling. 

\section{Related work}
\label{related}

\paragraph{In-context learning}
In many prior works investigating in-context learning sensitivity to ordering and selected examples (see \autoref{app:background} for description of in-context learning), a fixed number of examples are used \citep{zhang-etal-2022-active, lu-etal-2022-fantastically, min-etal-2022-rethinking}, with common guidance from prior ablations stating performance plateaus around $k=4$ \citep{wang-etal-2024-learning} and $k=8$ examples \citep{min-etal-2022-rethinking, wu-etal-2023-self}. Recent work has looked at the how the number of examples impacts performance on chain-of-thought reasoning benchmarks \citep{chen-etal-2023-many} and suggested that fewer examples may be needed, yet it is otherwise unclear the effect of number of demonstrations on other tasks and prompting frameworks, particularly when controlling for order and selected examples.

\paragraph{Monte Carlo sampling}
Monte Carlo sampling has been widely adopted in the data valuation literature to provide unbiased approximations of the Shapley value \citep{pmlr-v97-ghorbani19c}, based on prior work on Monte Carlo methods for Shapley value approximation \citep{mann1962values, castro2009polynomial, maleki2013bounding} (see \autoref{app:background} for additional details). Recent works have also applied principles from data valuation to in-context learning example selection and ordering \citep{guo-etal-2024-makes, chang-jia-2023-data, nguyen2023context}. While in data valuation, Monte Carlo sampling methods are used to calculate the marginal contribution of each data point averaged over a number of permutations, our motivation differs. In our setting, we are motivated by the utility of Monte Carlo sampling to provide an unbiased estimate of the influence of number of examples on in-context learning performance by reducing influence of ordering and selected examples.

\section{Method}\label{sec:method}
\begin{algorithm}[h]
  \caption{\textbf{Monte Carlo Sampling Method}}
  \label{alg:mcs}
      \begin{algorithmic}[1]
        \STATE \textbf{Input:} Training data $D_{trn}=\{1,...,n\}$, evaluation data $D_{tst}=\{1,...,m\}$, LLM \\ $\mathcal{M}$, performance metric $V_\mathcal{M}$, parameters: \\ $K$ (\# examples), $P$ (\# permutations)
        \STATE {\textbf{Output:} Average performance $\mu(V_\mathcal{M})$ for $k=\{0,...,K\}$} for one subset $S^K$
        \STATE Initialize $\mu_k = 0$ for $k=0,...,K$
        \STATE Randomly sample subset $S^K$ from $D_{trn}$
        \FOR{$p\in \{1,...,P\}$} 
            \STATE $\pi^p$: \text{Random permutation of} $S^K$
            \FOR{$k\in\{0,...,K\}$}
              \STATE $\mu_k \leftarrow \mu_k + V_{\mathcal{M}}(S[0:k]_{\pi})$
            \ENDFOR
        \ENDFOR
        \FOR{$k\in\{0,...,K\}$}
            \STATE $\mu_k \leftarrow \frac{1}{P}\mu_k$
        \ENDFOR
      \end{algorithmic}
\end{algorithm}
To study the effect of the number of examples, we aim to reduce the influence of ordering and selected examples as confounding factors. While averaging across trials in previous work helps reduce the influence of selected examples, it does not account for different orderings. Additionally, prior work on ordering only addresses up to $k=4$ \citep{lu-etal-2022-fantastically} as the possible permutations increase exponentially with respect to $k$. This motivates the use of Monte Carlo sampling with incremental exemplar additions as 1) the use of permutations reduces the influence of ordering, and 2) the averaging across multiple trials reduces the influence of selected examples. We explain this as well as our method in detail below.

Consider a training dataset $D_{trn}=\{(x_i ,y_i)\}_{i=1}^n$  that contains $n$ training instances. Given a fixed test set $D_{tst}$, we aim to draw $k$ exemplars from $D_{trn}$ and test model performance on $D_{tst}$. We let $S_\pi^k \subseteq D_{trn}$ denote a subset with some ordering $\pi$ and cardinality $k$, where $k \leq n$. Additionally, we let $V_\mathcal{M}$ denote the predictive performance of a model $\mathcal{M}$, e.g., prediction accuracy.

For each $k\in\{1,2,...,K\}$, the set of possible subsets $S^k=\{S_{(i)}^k\}_{i=1}^{_nC_k}$. The expected model performance for any given $S_{(i)}^k$ can be defined as:

\begin{figure*}
    \centering
    \begin{minipage}[t]{\linewidth}
        \begin{subfigure}{0.24\linewidth}
            \centering
            \includegraphics[width=\textwidth]{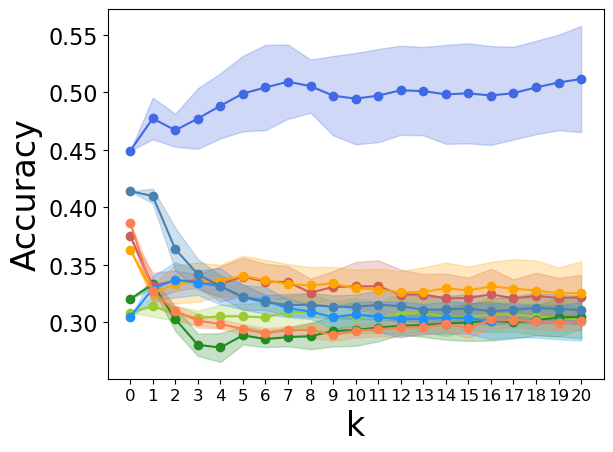}
            \caption{MNLI}\label{fig:removal-image1}
        \end{subfigure}%
        \hfill
        \begin{subfigure}{0.24\linewidth}
            \centering
            \includegraphics[width=\textwidth]{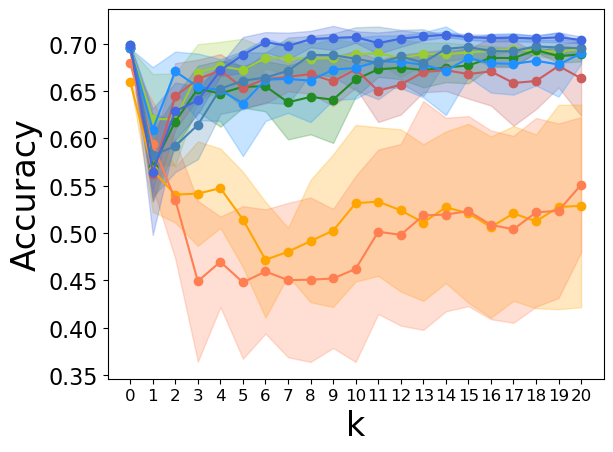}
            \caption{MRPC}\label{fig:removal-image2}
        \end{subfigure}
        \hfill
        \begin{subfigure}{0.24\linewidth}
            \centering
            \includegraphics[width=\textwidth]{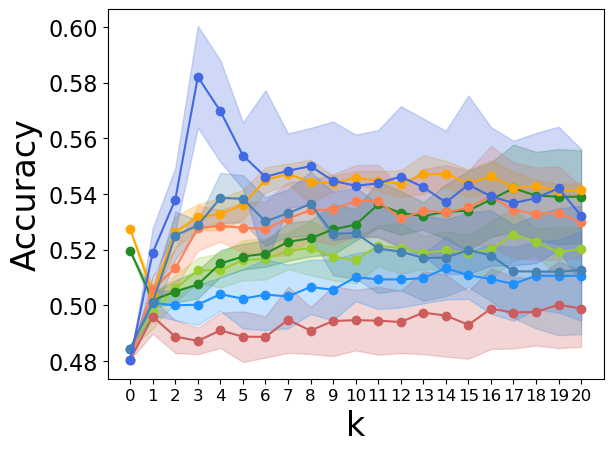}
            \caption{QNLI}\label{fig:removal-image3}
        \end{subfigure}
        \begin{subfigure}{0.24\linewidth}
            \centering
            \includegraphics[width=\textwidth]{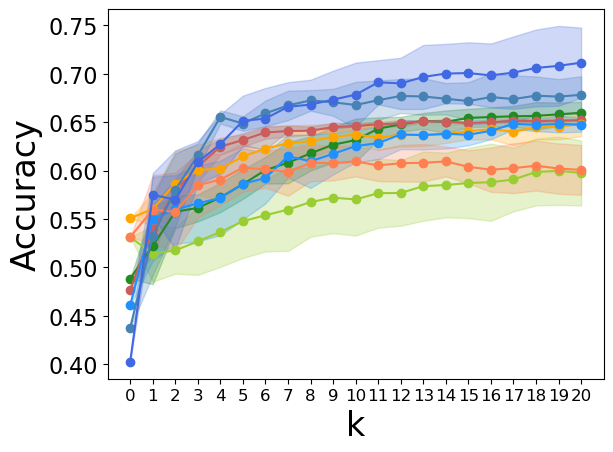}
            \caption{QQP}\label{fig:removal-image4}
        \end{subfigure}%
        \hfill
        \begin{subfigure}{0.24\linewidth}
            \centering
            \includegraphics[width=\textwidth]{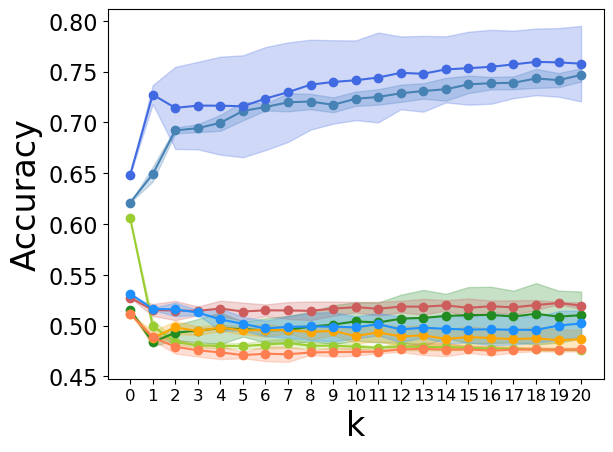}
            \caption{RTE}\label{fig:removal-image5}
        \end{subfigure}
        \hfill
            \begin{subfigure}{0.24\linewidth}
            \centering
            \includegraphics[width=\textwidth]{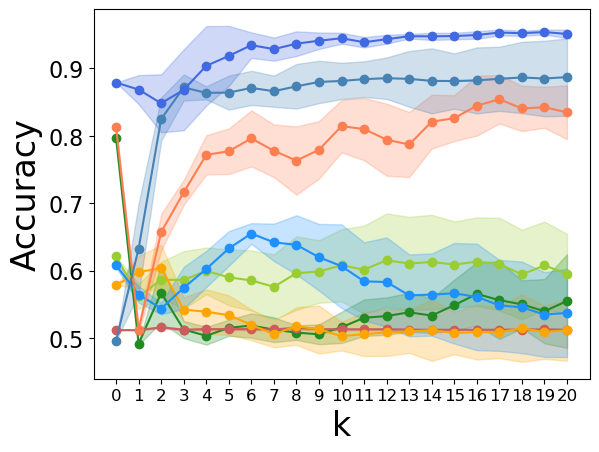}
            \caption{SST-2}\label{fig:removal-image6}
        \end{subfigure}
        \hfill
        \begin{subfigure}{0.24\linewidth}
            \centering
            \includegraphics[width=\textwidth]{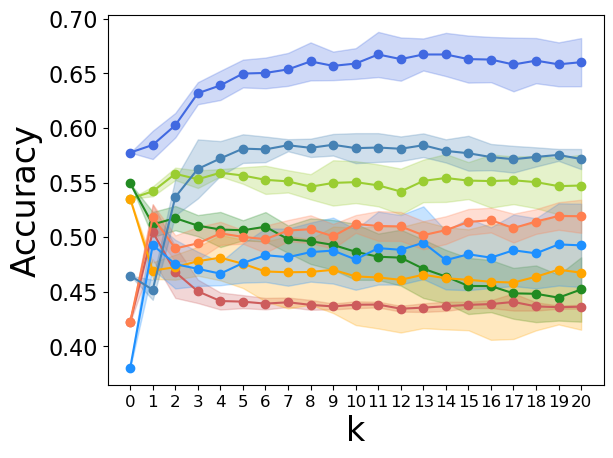}
            \caption{WNLI}\label{fig:removal-image7}
        \end{subfigure}%
        \hfill
        \begin{subfigure}{0.24\linewidth}
            \centering
            \includegraphics[width=\textwidth]{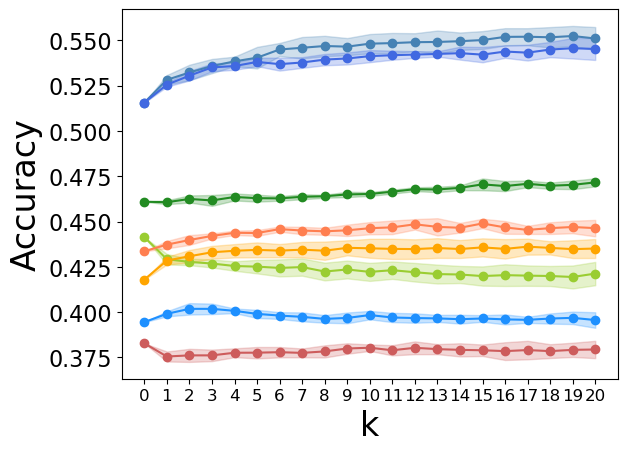}
            \caption{Hellaswag}\label{fig:removal-image8}
        \end{subfigure}
        \hfill
        \begin{subfigure}{\linewidth}
            \centering
            \vspace*{1mm}
            \includegraphics[width=0.7\textwidth]{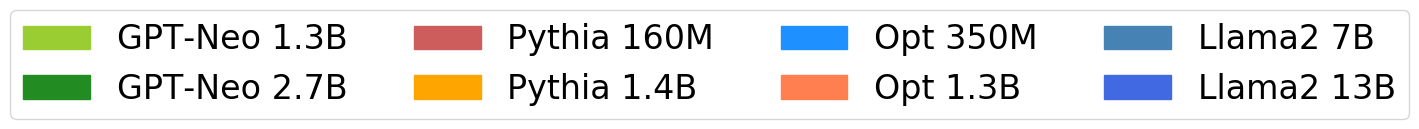}
        \end{subfigure}
        \hfill
            \begin{minipage}{.1cm}
            \vfill
            \end{minipage}
    \end{minipage}%
    \hfill
    \vspace*{-4mm}
    \begin{minipage}[c]{\linewidth}
        \caption{\label{fig:dataset-accs}In-context performance for each dataset and model. Results show the average of 20 permutations at each step $k$ in the proposed Monte Carlo sampling method. Shaded regions show standard deviation of 5 trials.}
    \end{minipage}
\end{figure*}

\begin{equation}\label{eq:exp}
    \mathbb{E}_{\pi\sim\Pi}[V_\mathcal{M}(S_{\pi(i)}^k)]
\end{equation}

\noindent where $V_\mathcal{M}(S_\pi^k)$ is the performance of model $\mathcal{M}$ using $S_\pi^k$ and $\Pi$ is the uniform distribution over all $k!$ permutations of $S_{(i)}^k$.

For each $k$, exhaustively permuting all $_nC_k$ subsets $S^k$, and averaging over the expected values would give us the expected value for each number of exemplars. In practice, however, this is computationally infeasible. Instead, we utilize Monte Carlo sampling to approximate this value. 

Specifically, we utilize the Monte Carlo sampling method adapted from \citet{pmlr-v97-ghorbani19c}. First, we define $K$ as the maximum exemplars we aim to use. Next, we sample $S^K$ from $D_{trn}$. For a predefined number of permutations of $S^K$, we iteratively scan from the first element to the final element, computing the model performance at each step. We average the performance for each $k$ across $p$ permutations. We provide the pseudo-code in \autoref{alg:mcs}. In practice, we perform this procedure over multiple trials, resampling $S^K$ for each trial and averaging over the result. In our experiments, we use $5$ trials and $p=20$ permutations. 

In addition to the computational speedup achievable by limiting the number of permutations, the primary benefit of our approach is that is circumvents the need to resample at each $k$. As we cannot exhaustively permute each $S^k$, if we resample for each $k$, there are $\frac{n!}{(n-k-1)!}$ possible permutations preceding $k$, each of which could produce differing downstream performance $V_\mathcal{M}(S_{\pi_{k-1}}^{k-1})$ due to the impact of $S^{k-1}$ and $\pi_{k-1}$ (selected examples and ordering, respectively). By sampling in this manner, we effectively eliminate the influence of $k-1$ in understanding performance at $k$.

\section{Experiment setup}

Detailed descriptions of our setup can be found in \autoref{app:setup}.

\paragraph{Models} We experiment with 8 models in total, listed in \autoref{table:models}. The selected models vary in size from 160M to 13B and represent four distinct families: Pythia \citep{biderman2023pythia}, OPT \citep{zhang2022opt}, GPT-Neo \citep{gpt-neo}, and Llama2 \citep{touvron2023llama}. 
 
\paragraph{Datasets} We use 8 datasets in our experiments across a diverse range of tasks previously represented in in-context learning analysis. Specifically, we perform experiments on natural language inference (MNLI, QNLI, WNLI), sentiment analysis (SST-2), commonsense reasoning (Hellaswag), paraphrase detection (MRPC, QQP), and textual entailment (RTE). 

\paragraph{Sampling and Aggregation Procedure} For each trial, we randomly sample $20$ in-context examples from the training set. We perform $20$ Monte Carlo permutations. Within each permutation, we iterate from the first example to the last, computing the accuracy on the validation set at each step. Using this procedure, we perform $5$ trials for each experiment. In aggregate, we perform $100$ permutations with each model and dataset combination. 

\section{Results}\label{sec:results}

\subsection{Analyzing Number of Examples}\label{subsec:number}
We plot the performance of each model across $k=\{1,2,...,20\}$ examples, where each step represents the addition of one in-context example from the Monte Carlo permutation. In line with common practice of reporting across multiple trials with standard deviation information, our results are averaged over 5 trials, controlling for selected examples and ordering effects. Results are reported in \autoref{fig:dataset-accs}.

Observed from the averaged results, our results align with ablations in prior work \citep{wu-etal-2023-self, min-etal-2022-rethinking}.
Specifically, much of the performance improvement across models and datasets occurs within the first $~8$ in-context examples, where performance then begins to plateau and only marginally increase. At face value, this result shows that prior recommendations with respect to the number of examples to use are not influenced by ordering effects as we had hypothesized prior to designing our sampling method.

\begin{figure}[t!]
    \centering
    \begin{minipage}[t]{\linewidth}
        \begin{subfigure}{0.49\linewidth}
            \centering
            \includegraphics[width=\textwidth]{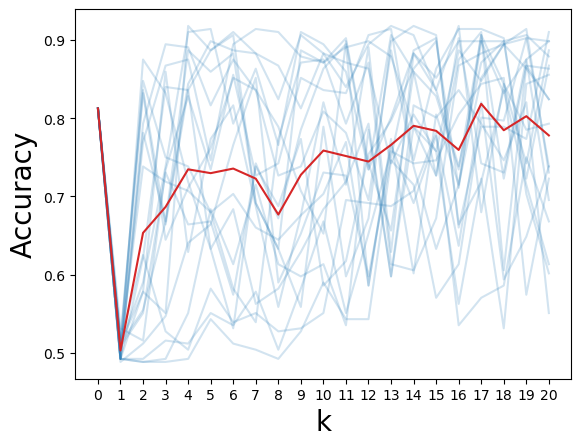}
            \caption{Opt-1.3B}\label{fig:removal-image3}
        \end{subfigure}
        \begin{subfigure}{0.49\linewidth}
            \centering
            \includegraphics[width=\textwidth]{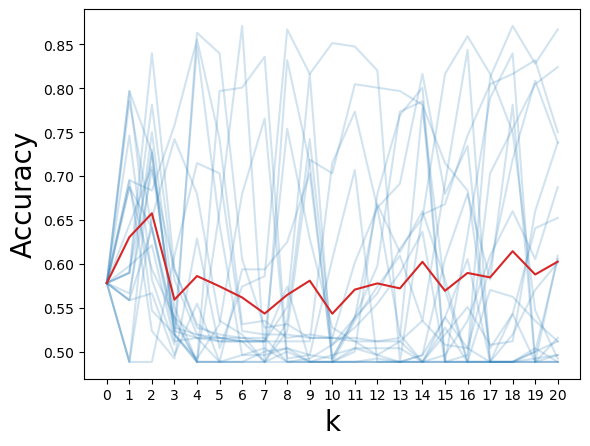}
            \caption{Pythia-1.4B}\label{fig:removal-image5}
        \end{subfigure}
        \hfill
            \begin{subfigure}{0.49\linewidth}
            \centering
            \includegraphics[width=\textwidth]{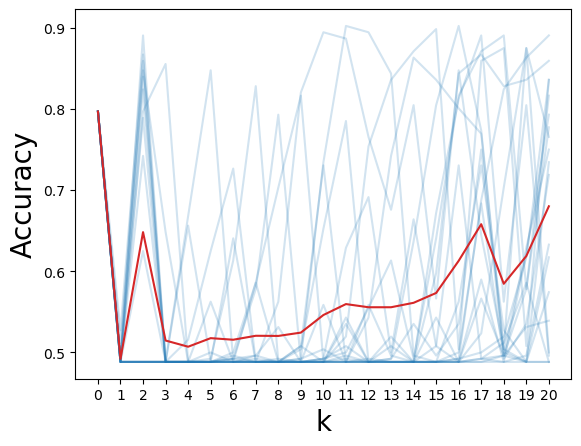}
            \caption{GPT-Neo-2.7B}\label{fig:removal-image6}
        \end{subfigure}
        \hfill
        \begin{subfigure}{0.49\linewidth}
            \centering
            \includegraphics[width=\textwidth]{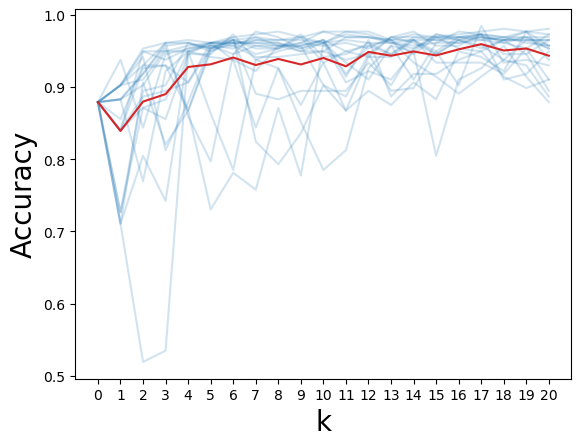}
            \caption{Llama2-13B}\label{fig:removal-image8}
        \end{subfigure}
        \hfill
            \begin{minipage}{.1cm}
            \vfill
            \end{minipage}
    \end{minipage}%
    \hfill
    \vspace*{-4mm}
    \begin{minipage}[c]{\linewidth}
        \caption{\label{fig:ovsst2} Results on SST-2. Blue lines represent individual permutations and red line indicates average across all permutations within one trial.}
    \end{minipage}
\vspace*{-4mm}
\end{figure}

\paragraph{Does performance consistently plateau at set numbers of exemplars?}
When viewing the results from individual permutations, we find that the performance plateaus by $k=4$ \citep{wang-etal-2024-learning} and $k=8$ \citep{min-etal-2022-rethinking, wu-etal-2023-self} examples are not observable within individual permutations. On the contrary, we continue to observe erratic performance changes up through $k=20$ examples. This suggests the previously observed plateaus on averaged results, both in prior work and recreated in \autoref{fig:dataset-accs}, may mask significant performance fluctuations, and the best performance within a selected example set may occur anywhere from $k=\{1, 2, ... 20\}$. Additionally, we observe that in some permutations, there is a performance drop at $k=1$. We investigate this further in the following section.

\paragraph{Is one example better than none?}\label{subsec:oneex}

Prior work has suggested that one demonstration outperforms no demonstrations even when a random label is used \citep{min-etal-2022-rethinking}. Other work has suggested some dependence on the dataset and model \citep{xie2022an, brown2020language}.
However, our results indicate that the performance of zero-shot vs. few-shot settings may be dependent on the selected example, regardless of dataset and model. To illustrate this, in \autoref{fig:one-point}, we plot the one-shot performance of each permutation across all 5 trials on the MNLI dataset using each model, with the zero-shot performance of the model indicated in red. 

In contradiction to prior work, our results across nearly all models and datasets (see \autoref{app:results}) indicate that performance in one-shot settings can vary between significant performance improvements and significant performance degradation, depending on the selected example.
This contradiction raises the question of what qualities of in-context examples can lead to such significant performance degradation in one-shot settings, and whether these have any impact when used within a $k$-shot setting. 

Our analysis in one-shot settings across different permutations indicates that one-shot performance is more sensitive to the selected example than previously thought.

\begin{figure}[t!]
    \centering
    \begin{minipage}[t]{\linewidth}
        \begin{subfigure}{0.49\linewidth}
            \centering
            \includegraphics[width=\textwidth]{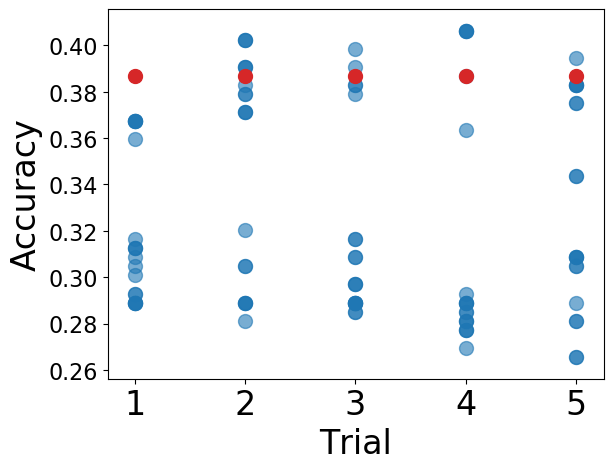}
            \caption{Opt-1.3B}\label{fig:removal-image1}
        \end{subfigure}%
        \hfill
        \begin{subfigure}{0.49\linewidth}
            \centering
            \includegraphics[width=\textwidth]{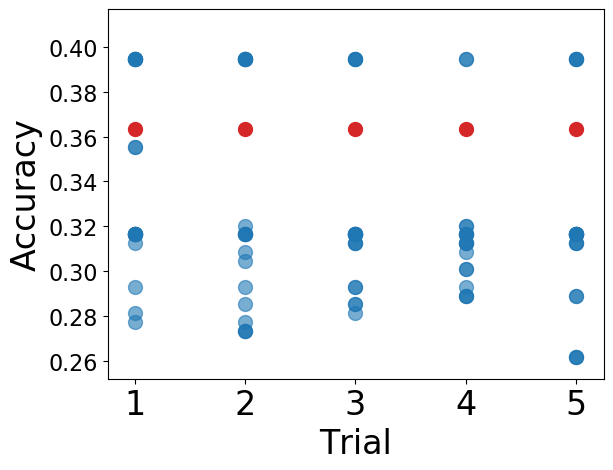}
            \caption{Pythia-1.4B}\label{fig:removal-image2}
        \end{subfigure}
        \hfill
        \begin{subfigure}{0.49\linewidth}
            \centering
            \includegraphics[width=\textwidth]{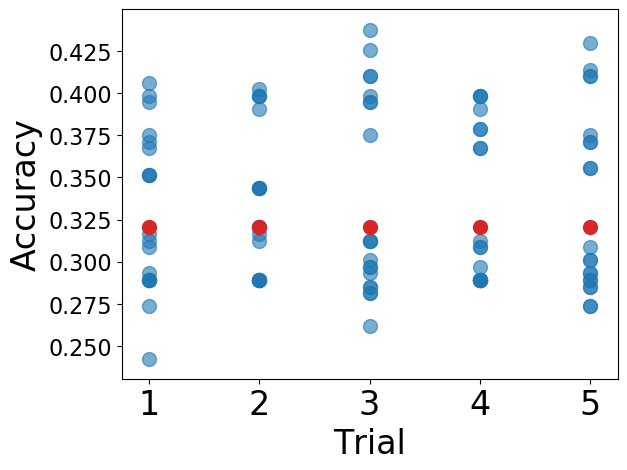}
            \caption{GPT-Neo-2.7B}\label{fig:removal-image3}
        \end{subfigure}
        \hfill
        \begin{subfigure}{0.49\linewidth}
            \centering
            \includegraphics[width=\textwidth]{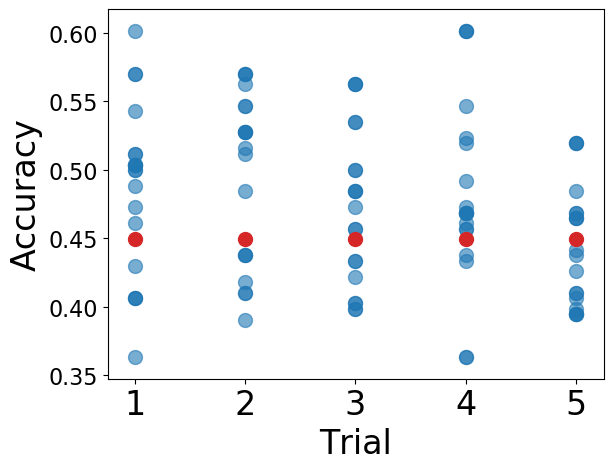}
            \caption{Llama2-13B}\label{fig:removal-image4}
        \end{subfigure}%
    \end{minipage}%
    \hfill
    \begin{minipage}[c]{\linewidth}
        \caption{\label{fig:one-point}One-shot MNLI performance across 5 trials. Each blue point represents the accuracy using the first exemplar in a permutation. 
        Red points indicate zero-shot performance. Results show that zero-shot settings can outperform one-shot settings, dependent upon the selected example.}
    \end{minipage}
\end{figure}

\subsection{Exemplar Selection}\label{sec:select}
\begin{figure}[t]
    \centering
    \includegraphics[width=0.5\textwidth]{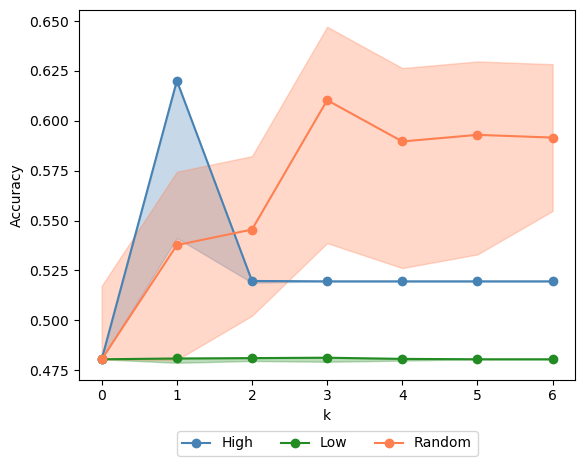}
    \caption{\label{fig:bwr} Performance with Llama2-13B on QNLI dataset, using in-context subsets containing the highest-performing and lowest-performing data points on average from \autoref{subsec:number}, along with a random baseline. Results represent 20 permutations, with standard deviation displayed as the shaded region for each line.}
\end{figure}
We are interested in whether we can apply our sampling method to selecting in-context examples. 
Using our sampling method, each exemplar appears at different $k$'s under different permutations. Therefore, we are effectively averaging out different orderings and $k$'s. It follows that we would expect exemplars associated with high average accuracy within different permutations to be associated with higher model performance overall, while being robust across orderings and different $k$'s
Further, we would expect exemplars with low average accuracies to consistently lead to poor performance.

Given the token limitations of context windows and computational time associated with increasing $k$, to increase the example candidate set, we use $Z$-score as a means to compare examples across trials. This allows us to limit context to $k=20$ while increasing the overall candidate set and accounting for the impact of using different selected examples in each trial. We perform the sampling procedure described in \autoref{sec:method} with Llama2-13B on QNLI and perform the following calculation.

For a given trial $t$, we compute the average accuracy associated with each example $e$, denoted $\mu_e$. We then take the mean over all example averages in the trial, $\mu_t=\frac{1}{k}\Sigma_{e=1}^k \mu_e$ with standard deviation $\sigma_t=\sqrt{\frac{\Sigma_{e=1}^k(\mu_e-\mu_t)}{k}}$. Finally, we compute the $Z$-score for each example in the trial, where $Z=\frac{\mu_e-\mu_t}{\sigma_t}$
After computing scores for all trials, we identify points with a $Z$-score $Z>1\lor Z<1$, and select the $6$ examples with the highest and lowest scores as our \textit{high} and \textit{low} set, respectively. We report results using each set, as well as a random baseline for in-context learning in \autoref{fig:bwr} using the same sampling procedure as in \autoref{subsec:number}. 

When using points identified via Monte Carlo sampling as consistently high or low performing, we see a greater robustness to ordering sensitivity across $k$'s, as evidenced by the minimal variance exhibited by these subsets. This is contrasted with the random baseline which exhibits high variance across different $k$'s and orderings.

Interestingly, whereas our experiments in \autoref{subsec:oneex} showed numerous instances where one-shot performance was worse than zero-shot performance, we do not observe this occurring with the lowest average performing data points. Rather, we see performance maintained at a zero-shot level. Further, while high-performing examples consistently performs above zero-shot accuracy and exhibit greater stability over varying $k$ across permutations, the random selection exhibited the highest performance for $k>1$. This was an unexpected result: using our Monte Carlo-based selection method led to more robust performance across $k$ examples, but it resulted in an overall performance decrease compared to random sampling. 

Our results when selecting in-context examples that on average had high or low scores across Monte Carlo permutations indicate that subsets comprised of points at each end of the spectrum exhibit lower sensitivity to ordering and number of demonstrations. While this may be promising in terms of identifying methods to increase robustness of in-context learning performance, the random selection baseline exhibited both higher overall performance and higher variance across orderings and number of examples. This raises the question of whether there is an existing performance vs. robustness tradeoff, and of what qualities contribute to example deviation from the mean performance, as well as how methods can identify examples that possess these qualities that also lead to higher performance gains.

\section{Conclusion}
In this work, we proposed a Monte Carlo-based sampling method to investigate previous guidance regarding number of examples to use and one- vs. zero-shot settings. We further investigated whether this method could be used to select in-context examples, finding a performance vs. robustness tradeoff.

\section{Limitations}
Our results are reported on models up to 13B parameters due to constraints posed by our available computational resources. We acknowledge this as a limitation, however, as our results are consistent across the model sizes we utilized, we believe our results should generalize to larger models.

\section*{Acknowledgments}
We thank the reviewers for their helpful feedback. This research was supported in part by NSF III \#2007492.

\bibliography{custom.bib}

\clearpage

\appendix
\section{Additional Background}\label{app:background}

\paragraph{In-Context Learning: } In-context learning enables pre-trained models to learn an unseen task using a set of exemplars concatenated in the context window. Formally, given a test example $x$, in-context learning concatenates $K$ demonstration examples to the task instruction $I$, where $S = \{x_{\pi (i)}, y_{\pi i}\}_{i=1}^K$ denotes the example set given some order $\pi$.

\paragraph{Monte Carlo Sampling: } Within the context of data valuation, the underlying idea of Monte Carlo sampling is to sample random permutations of the data points and iterate from the first to last element in each permutation. 
Specifically, for $p$ Monte Carlo iterations, a dataset $D$ is randomly permuted. Following, these methods scan from the first element of the permutation to the last element of the permutation and compute the performance of the model at each timestep. In data valuation, Monte Carlo sampling methods are used to calculate the marginal contribution of each data point averaged over a number of permutations.

\section{Experiment Setup Details}\label{app:setup}

\begin{table}[t]
\centering
\begin{tabular}{ll}
  \toprule
  \textbf{Model} & \textbf{Parameters} \\
  \midrule
  \multirow{2}{*}{Pythia \citep{biderman2023pythia}} 
  & 160M \\
  & 1.4B \\ 
  \midrule
    \multirow{2}{*}{OPT \citep{zhang2022opt}} & 350M \\
  & 1.3B \\
  \midrule
  \multirow{2}{*}{GPT-Neo \citep{gpt-neo}} & 1.3B \\
  & 2.7B\\
  \midrule
  \multirow{2}{*}{LLaMa2 \citep{touvron2023llama}} & 7B \\
  & 13B \\
  \bottomrule
\end{tabular}
\caption{Models used in our experiments. We select a range of different model families and parameter sizes. Parameter range is upper bounded based on available compute. \label{table:models}}
\end{table}
\begin{table*}[t]
\centering
\begin{tabular}{lllll}
  \toprule
  \textbf{Dataset} & \textbf{Task} & \textbf{\#Train} & \textbf{\#Val} & \textbf{\#Classes} \\
  \midrule
  MNLI \citep{williams2018broad} & Natural Language Inference & 393k & 9.82k & 3\\
  MRPC \citep{dolan2004unsupervised} & Paraphrase Detection & 3.67k & 408 & 2 \\
  QNLI \citep{wang-etal-2018-glue} & Natural Language Inference & 105k & 5.46k & 2\\
  QQP \citep{Quora} & Paraphrase Detection & 364k & 40.4k & 2 \\
  RTE \tablefootnote{\citep{dagan2006pascal, bar2006second, giampiccolo2007third, bentivogli2009fifth}} & Textual Entailment & 2.49k & 277 & 2 \\
  SST-2 \citep{socher2013recursive} & Sentiment Analysis & 67.3k & 872 & 2 \\
  WNLI \citep{levesque2011winograd} & Natural Language Inference & 635 & 71 & 2 \\
  Hellaswag \citep{zellers2019hellaswag} & Commonsense Reasoning & 39.9k & 10k & 4 \\
  \bottomrule
\end{tabular}
\caption{\label{table:datasets} Datasets used in our experiments. We use the distributions available from Huggingface \citep{lhoest-etal-2021-datasets}, and use the respective validation sets to measure performance.}
\end{table*}

We adapt the Language Model Evaluation Harness package \citep{eval-harness} to conduct our experiments. All experiments use the package's default prompts for each dataset. 

Details on models and datasets used can be found in \autoref{table:models} and \autoref{table:datasets}, respectively. 

Notably, the upper bound of the parameter range for models is due to our resource constraint, as each experiment is run using a single NVIDIA A100 GPU. 

For each dataset, we utilize the splits available from Huggingface. As the GLUE benchmark datasets do not have labeled test sets, we use the validation sets for evaluation. Additionally, as we are performing inference after the addition of every example within each permutation, we follow a protocol from prior work and sub-sample 256 instances from the validation set to control inference overhead \citep{lu-etal-2022-fantastically}. 

\section{Additional results}\label{app:results}
This appendix contains:
\begin{itemize}
    \item Additional scatter plots (\autoref{subsec:oneex})
    \item Single trial averages overlaying individual permutations (\autoref{subsec:oneex})
\end{itemize}

\begin{figure*}[h]
    \centering
    \begin{minipage}[t]{\linewidth}
        \begin{subfigure}{0.24\linewidth}
            \centering
            \includegraphics[width=\textwidth]{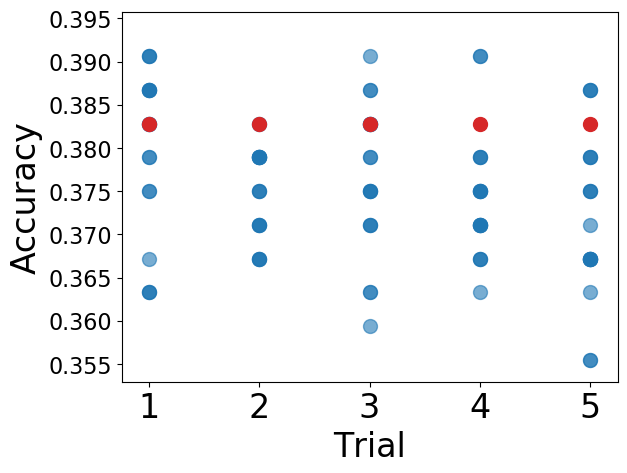}
            \caption{Pythia-160M}\label{fig:removal-image1}
        \end{subfigure}%
        \hfill
        \begin{subfigure}{0.24\linewidth}
            \centering
            \includegraphics[width=\textwidth]{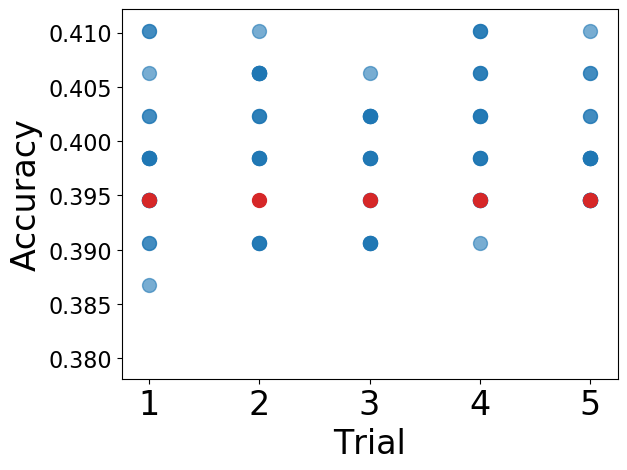}
            \caption{Opt-350M}\label{fig:removal-image2}
        \end{subfigure}
        \hfill
        \begin{subfigure}{0.24\linewidth}
            \centering
            \includegraphics[width=\textwidth]{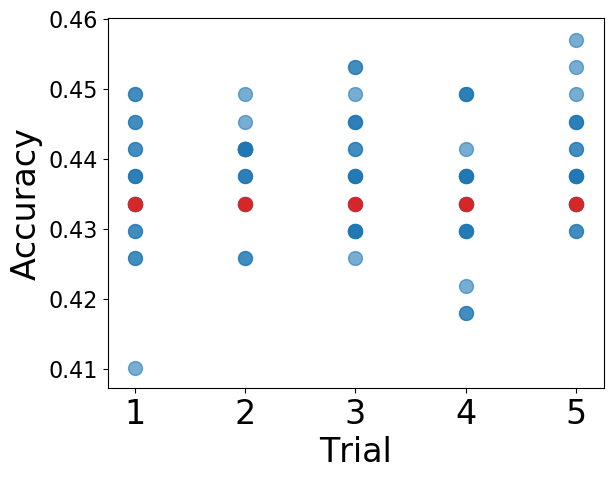}
            \caption{Opt-1.3B}\label{fig:removal-image3}
        \end{subfigure}
        \begin{subfigure}{0.24\linewidth}
            \centering
            \includegraphics[width=\textwidth]{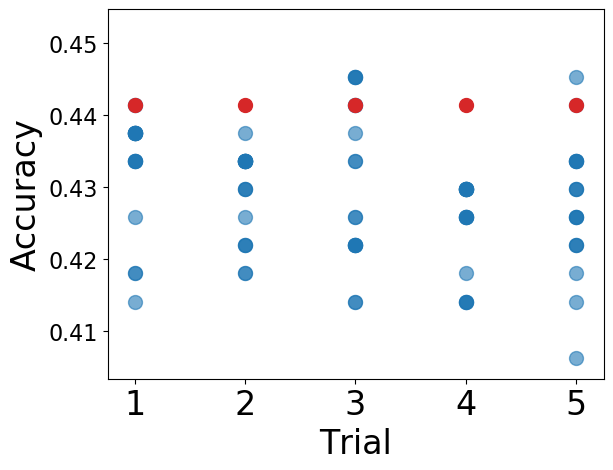}
            \caption{GPT-Neo-1.3B}\label{fig:removal-image4}
        \end{subfigure}%
        \hfill
        \begin{subfigure}{0.24\linewidth}
            \centering
            \includegraphics[width=\textwidth]{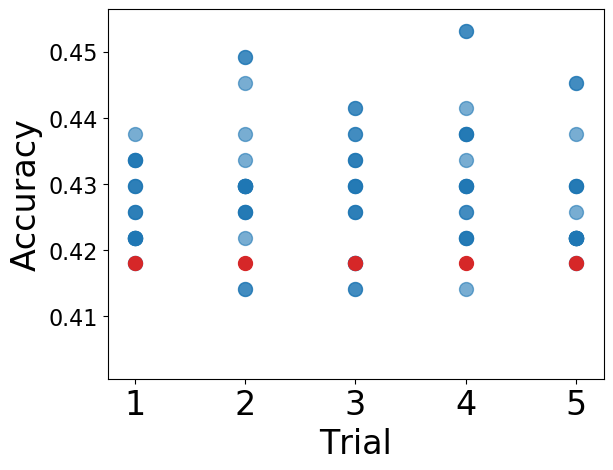}
            \caption{Pythia-1.4B}\label{fig:removal-image5}
        \end{subfigure}
        \hfill
            \begin{subfigure}{0.24\linewidth}
            \centering
            \includegraphics[width=\textwidth]{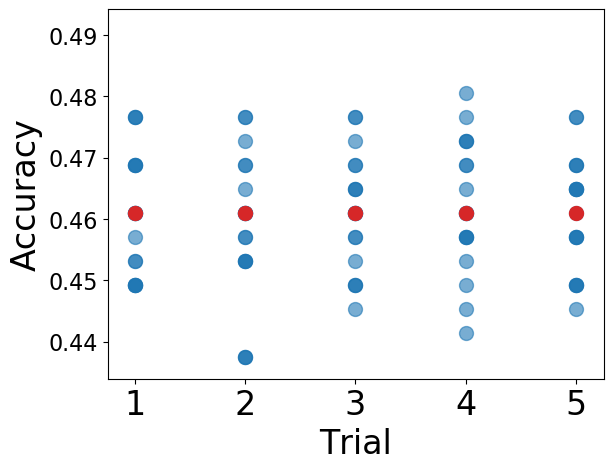}
            \caption{GPT-Neo-2.7B}\label{fig:removal-image6}
        \end{subfigure}
        \hfill
        \begin{subfigure}{0.24\linewidth}
            \centering
            \includegraphics[width=\textwidth]{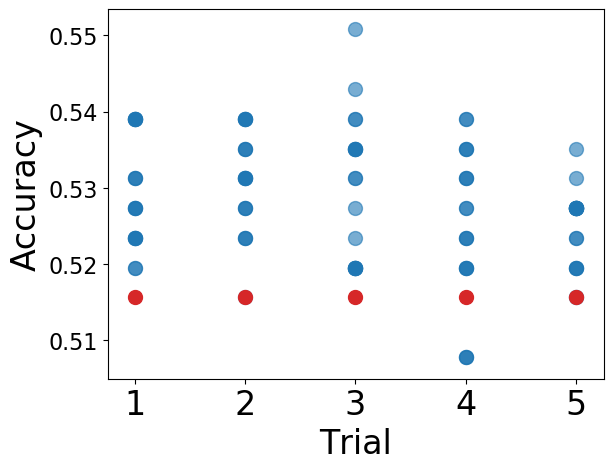}
            \caption{Llama2-7B}\label{fig:removal-image7}
        \end{subfigure}%
        \hfill
        \begin{subfigure}{0.24\linewidth}
            \centering
            \includegraphics[width=\textwidth]{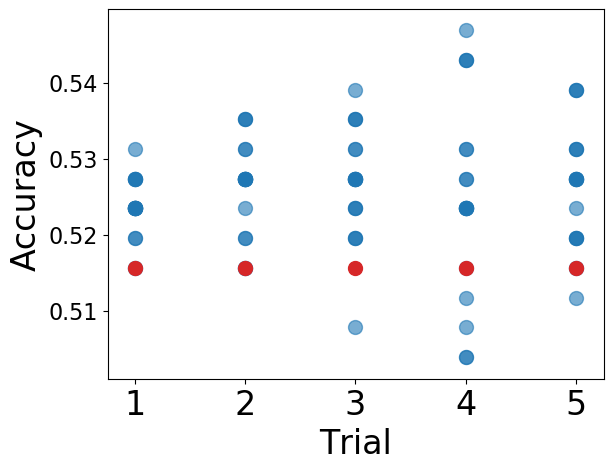}
            \caption{Llama2-13B}\label{fig:removal-image8}
        \end{subfigure}
        \hfill
            \begin{minipage}{.1cm}
            \vfill
            \end{minipage}
    \end{minipage}%
    \hfill
    \begin{minipage}[c]{\linewidth}
        \caption{\label{fig:}One-shot in-context learning performance on the Hellaswag dataset across 5 trials.}
    \end{minipage}
\end{figure*}
\begin{figure*}
    \centering
    \begin{minipage}[t]{\linewidth}
        \begin{subfigure}{0.24\linewidth}
            \centering
            \includegraphics[width=\textwidth]{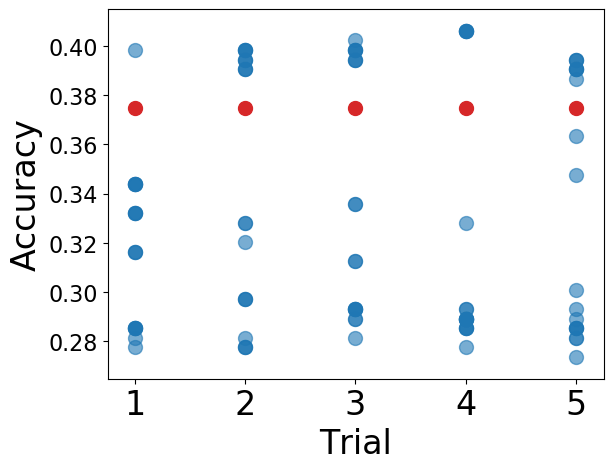}
            \caption{Pythia-160M}\label{fig:removal-image1}
        \end{subfigure}%
        \hfill
        \begin{subfigure}{0.24\linewidth}
            \centering
            \includegraphics[width=\textwidth]{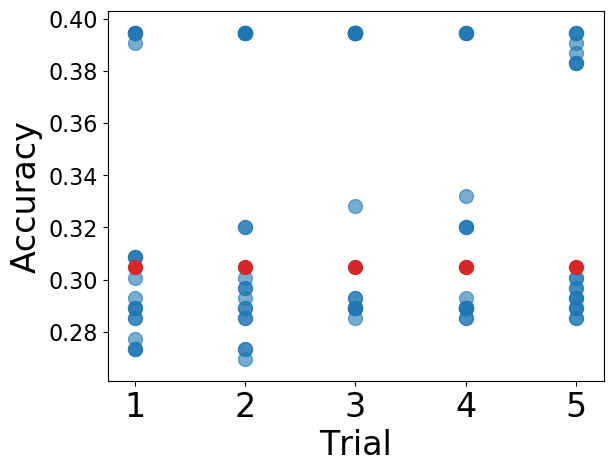}
            \caption{Opt-350M}\label{fig:removal-image2}
        \end{subfigure}
        \hfill
        \begin{subfigure}{0.24\linewidth}
            \centering
            \includegraphics[width=\textwidth]{figures/mnli-opt-1.3b-sc-tight.png}
            \caption{Opt-1.3B}\label{fig:removal-image3}
        \end{subfigure}
        \begin{subfigure}{0.24\linewidth}
            \centering
            \includegraphics[width=\textwidth]{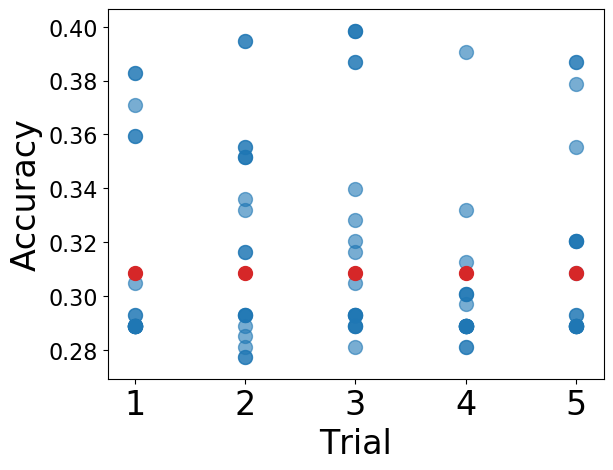}
            \caption{GPT-Neo-1.3B}\label{fig:removal-image4}
        \end{subfigure}%
        \hfill
        \begin{subfigure}{0.24\linewidth}
            \centering
            \includegraphics[width=\textwidth]{figures/mnli-pythia-1.4b-sc-tight.png}
            \caption{Pythia-1.4B}\label{fig:removal-image5}
        \end{subfigure}
        \hfill
            \begin{subfigure}{0.24\linewidth}
            \centering
            \includegraphics[width=\textwidth]{figures/mnli-gpt-neo-2.7b-sc-tight.png}
            \caption{GPT-Neo-2.7B}\label{fig:removal-image6}
        \end{subfigure}
        \hfill
        \begin{subfigure}{0.24\linewidth}
            \centering
            \includegraphics[width=\textwidth]{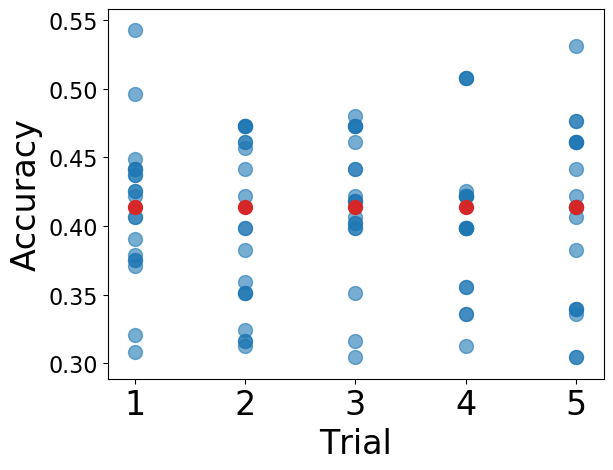}
            \caption{Llama2-7B}\label{fig:removal-image7}
        \end{subfigure}%
        \hfill
        \begin{subfigure}{0.24\linewidth}
            \centering
            \includegraphics[width=\textwidth]{figures/mnli-llama2-13b-sc-tight.png}
            \caption{Llama2-13B}\label{fig:removal-image8}
        \end{subfigure}
        \hfill
            \begin{minipage}{.1cm}
            \vfill
            \end{minipage}
    \end{minipage}%
    \hfill
    \begin{minipage}[c]{\linewidth}
        \caption{\label{fig:one-point}One-shot in-context learning performance on the MNLI dataset across 5 trials. Each blue point represents the accuracy using the first in-context example of a permutation, with $20$ permutations per trial. The red points indicate the zero-shot performance of the model. Results indicate that zero-shot settings can outperform one-shot settings, dependent upon the selected in-context example.}
    \end{minipage}
\end{figure*}
\begin{figure*}[h]
    \centering
    \begin{minipage}[t]{\linewidth}
        \begin{subfigure}{0.24\linewidth}
            \centering
            \includegraphics[width=\textwidth]{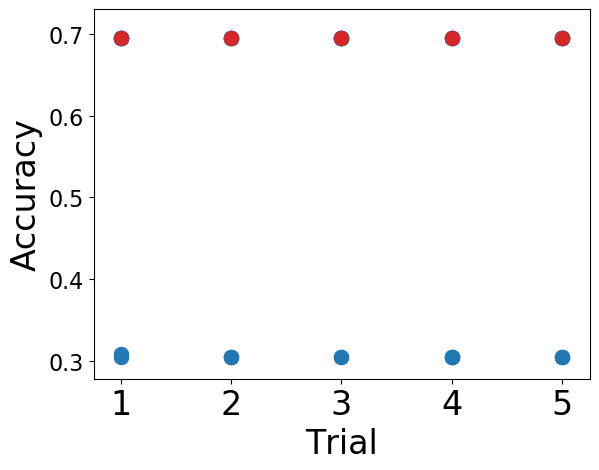}
            \caption{Pythia-160M}\label{fig:removal-image1}
        \end{subfigure}%
        \hfill
        \begin{subfigure}{0.24\linewidth}
            \centering
            \includegraphics[width=\textwidth]{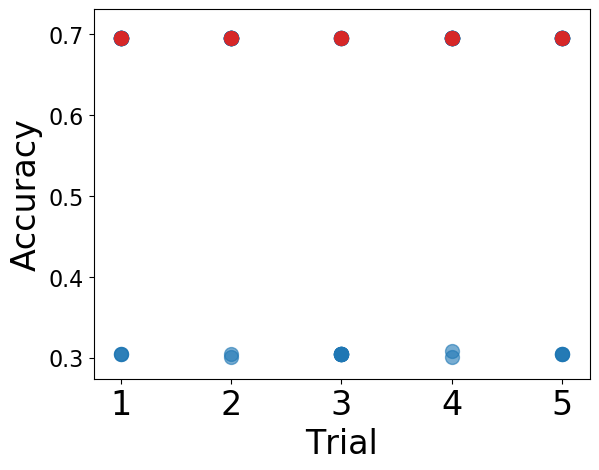}
            \caption{Opt-350M}\label{fig:removal-image2}
        \end{subfigure}
        \hfill
        \begin{subfigure}{0.24\linewidth}
            \centering
            \includegraphics[width=\textwidth]{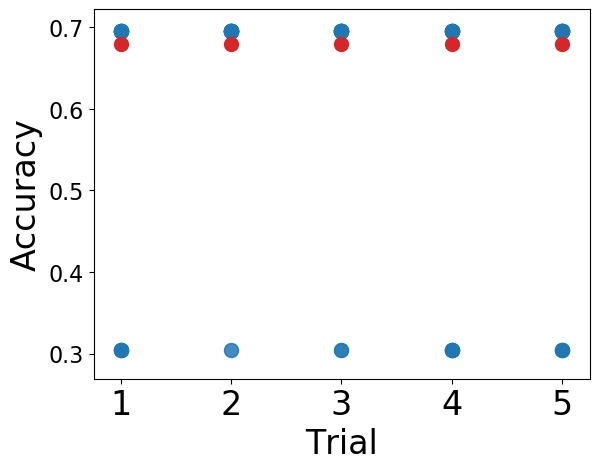}
            \caption{Opt-1.3B}\label{fig:removal-image3}
        \end{subfigure}
        \begin{subfigure}{0.24\linewidth}
            \centering
            \includegraphics[width=\textwidth]{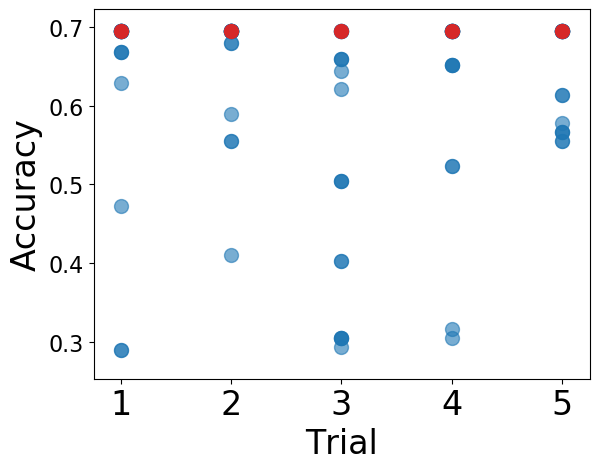}
            \caption{GPT-Neo-1.3B}\label{fig:removal-image4}
        \end{subfigure}%
        \hfill
        \begin{subfigure}{0.24\linewidth}
            \centering
            \includegraphics[width=\textwidth]{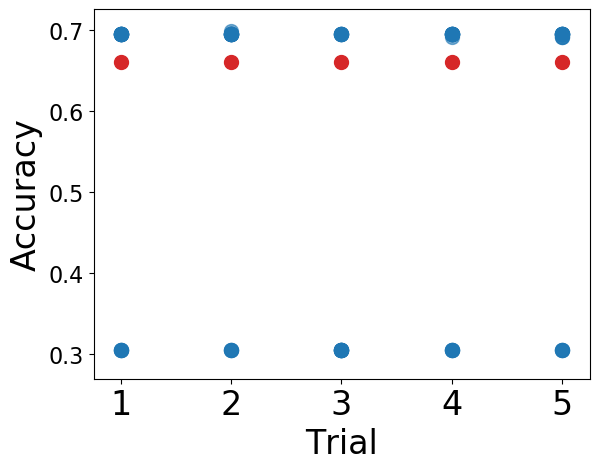}
            \caption{Pythia-1.4B}\label{fig:removal-image5}
        \end{subfigure}
        \hfill
            \begin{subfigure}{0.24\linewidth}
            \centering
            \includegraphics[width=\textwidth]{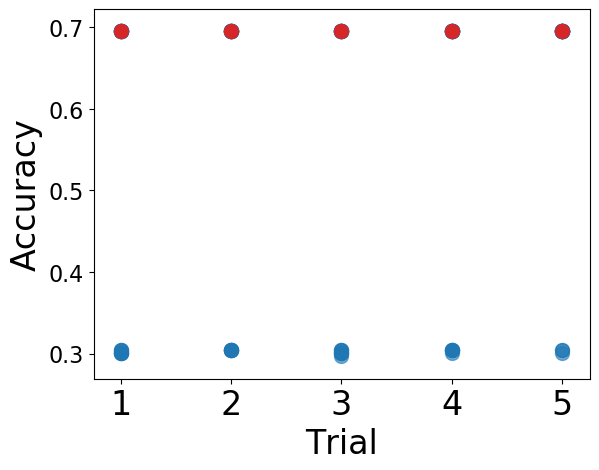}
            \caption{GPT-Neo-2.7B}\label{fig:removal-image6}
        \end{subfigure}
        \hfill
        \begin{subfigure}{0.24\linewidth}
            \centering
            \includegraphics[width=\textwidth]{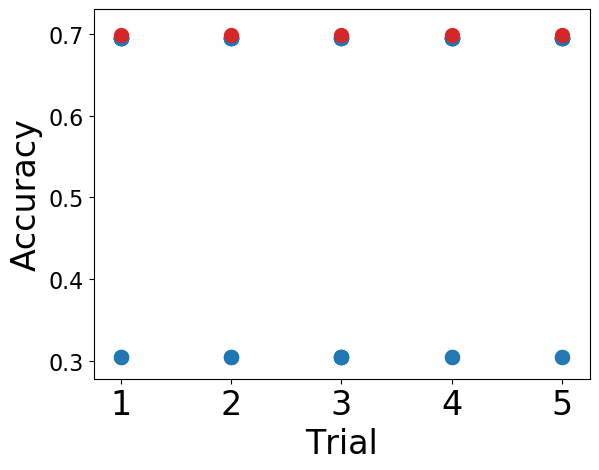}
            \caption{Llama2-7B}\label{fig:removal-image7}
        \end{subfigure}%
        \hfill
        \begin{subfigure}{0.24\linewidth}
            \centering
            \includegraphics[width=\textwidth]{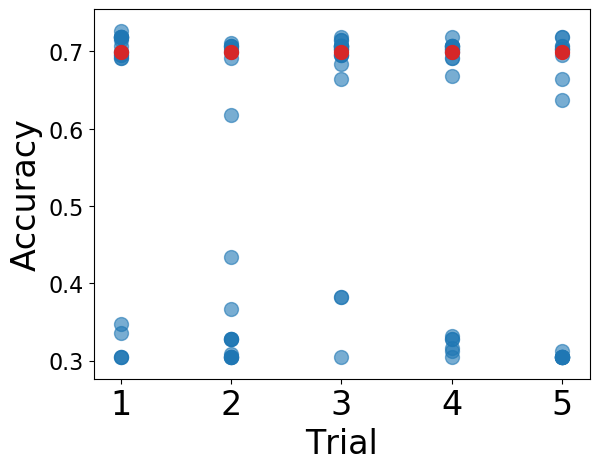}
            \caption{Llama2-13B}\label{fig:removal-image8}
        \end{subfigure}
        \hfill
            \begin{minipage}{.1cm}
            \vfill
            \end{minipage}
    \end{minipage}%
    \hfill
    \begin{minipage}[c]{\linewidth}
        \caption{\label{fig:}One-shot in-context learning performance on the MRPC dataset across 5 trials.}
    \end{minipage}
\end{figure*}
\begin{figure*}[h]
    \centering
    \begin{minipage}[t]{\linewidth}
        \begin{subfigure}{0.24\linewidth}
            \centering
            \includegraphics[width=\textwidth]{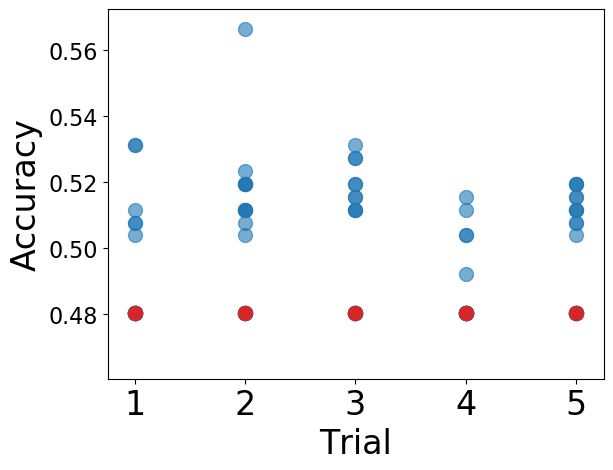}
            \caption{Pythia-160M}\label{fig:removal-image1}
        \end{subfigure}%
        \hfill
        \begin{subfigure}{0.24\linewidth}
            \centering
            \includegraphics[width=\textwidth]{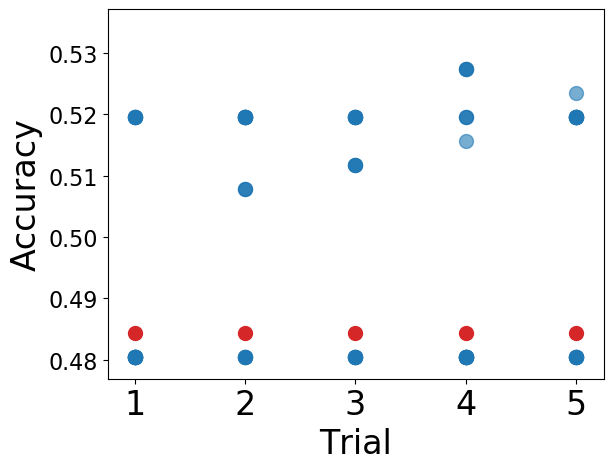}
            \caption{Opt-350M}\label{fig:removal-image2}
        \end{subfigure}
        \hfill
        \begin{subfigure}{0.24\linewidth}
            \centering
            \includegraphics[width=\textwidth]{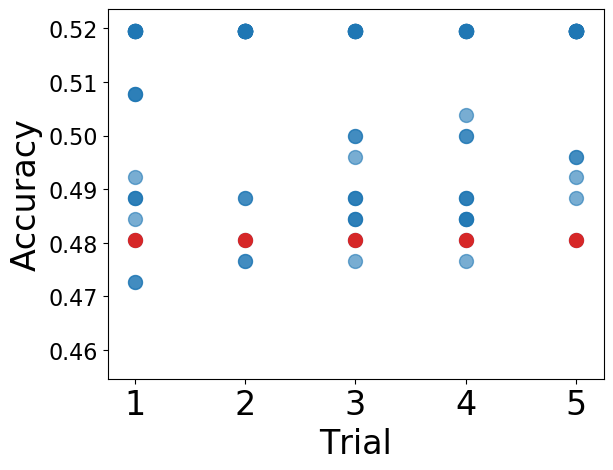}
            \caption{Opt-1.3B}\label{fig:removal-image3}
        \end{subfigure}
        \begin{subfigure}{0.24\linewidth}
            \centering
            \includegraphics[width=\textwidth]{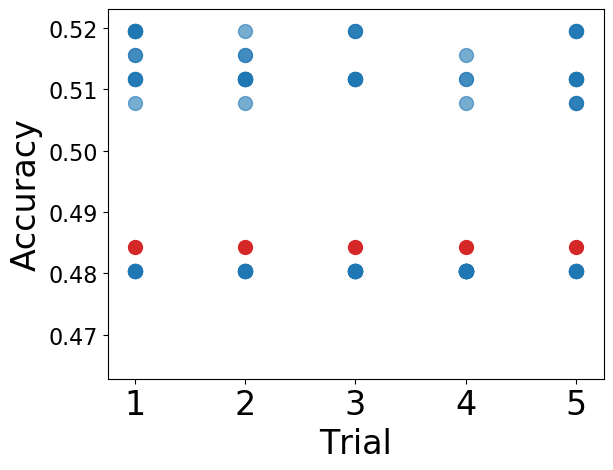}
            \caption{GPT-Neo-1.3B}\label{fig:removal-image4}
        \end{subfigure}%
        \hfill
        \begin{subfigure}{0.24\linewidth}
            \centering
            \includegraphics[width=\textwidth]{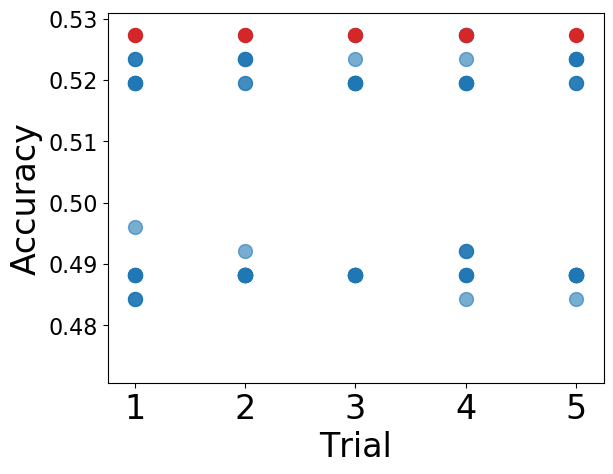}
            \caption{Pythia-1.4B}\label{fig:removal-image5}
        \end{subfigure}
        \hfill
            \begin{subfigure}{0.24\linewidth}
            \centering
            \includegraphics[width=\textwidth]{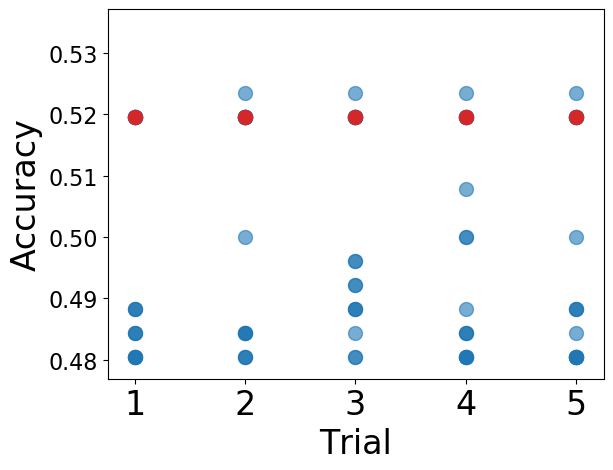}
            \caption{GPT-Neo-2.7B}\label{fig:removal-image6}
        \end{subfigure}
        \hfill
        \begin{subfigure}{0.24\linewidth}
            \centering
            \includegraphics[width=\textwidth]{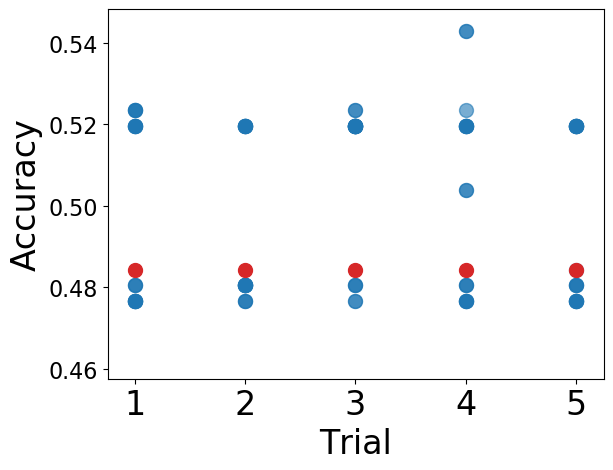}
            \caption{Llama2-7B}\label{fig:removal-image7}
        \end{subfigure}%
        \hfill
        \begin{subfigure}{0.24\linewidth}
            \centering
            \includegraphics[width=\textwidth]{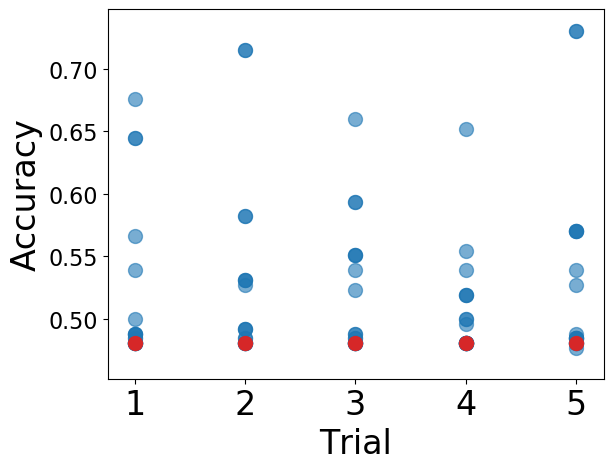}
            \caption{Llama2-13B}\label{fig:removal-image8}
        \end{subfigure}
        \hfill
            \begin{minipage}{.1cm}
            \vfill
            \end{minipage}
    \end{minipage}%
    \hfill
    \begin{minipage}[c]{\linewidth}
        \caption{\label{fig:}One-shot in-context learning performance on the QNLI dataset across 5 trials.}
    \end{minipage}
\end{figure*}
\begin{figure*}[h]
    \centering
    \begin{minipage}[t]{\linewidth}
        \begin{subfigure}{0.24\linewidth}
            \centering
            \includegraphics[width=\textwidth]{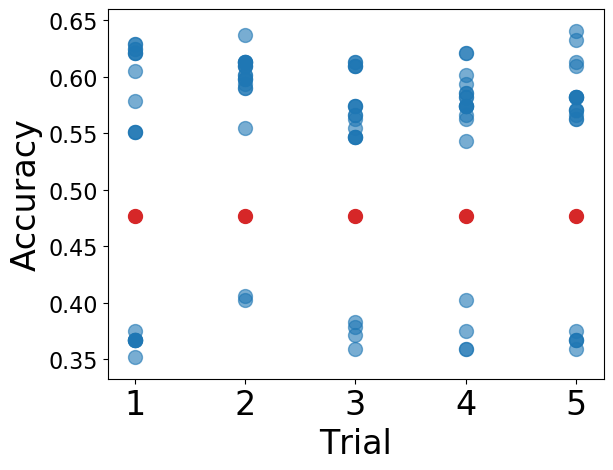}
            \caption{Pythia-160M}\label{fig:removal-image1}
        \end{subfigure}%
        \hfill
        \begin{subfigure}{0.24\linewidth}
            \centering
            \includegraphics[width=\textwidth]{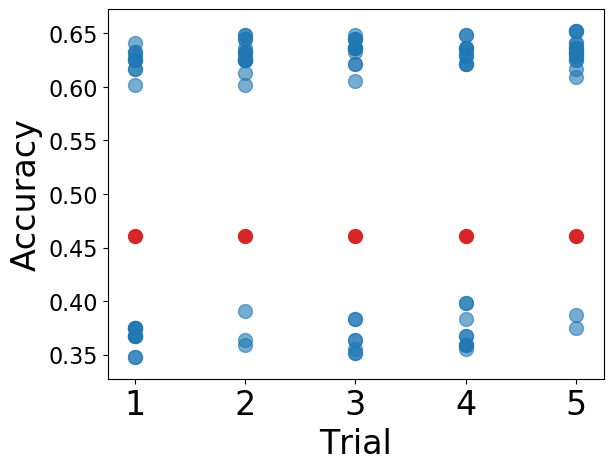}
            \caption{Opt-350M}\label{fig:removal-image2}
        \end{subfigure}
        \hfill
        \begin{subfigure}{0.24\linewidth}
            \centering
            \includegraphics[width=\textwidth]{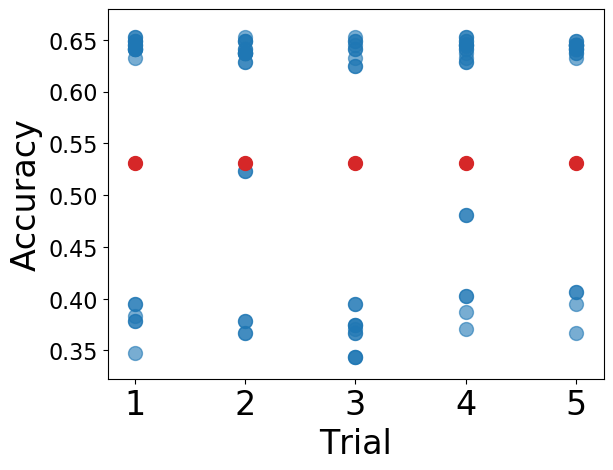}
            \caption{Opt-1.3B}\label{fig:removal-image3}
        \end{subfigure}
        \begin{subfigure}{0.24\linewidth}
            \centering
            \includegraphics[width=\textwidth]{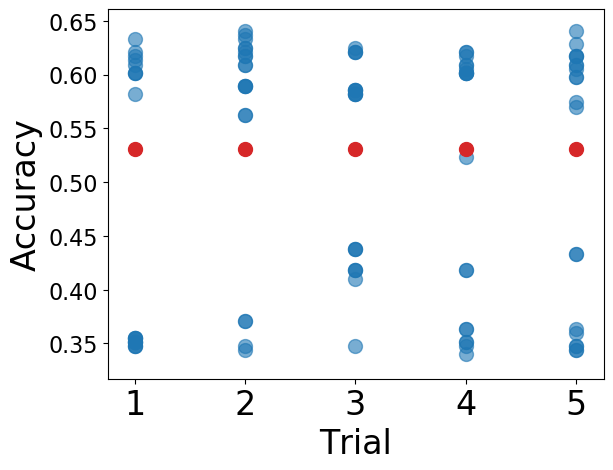}
            \caption{GPT-Neo-1.3B}\label{fig:removal-image4}
        \end{subfigure}%
        \hfill
        \begin{subfigure}{0.24\linewidth}
            \centering
            \includegraphics[width=\textwidth]{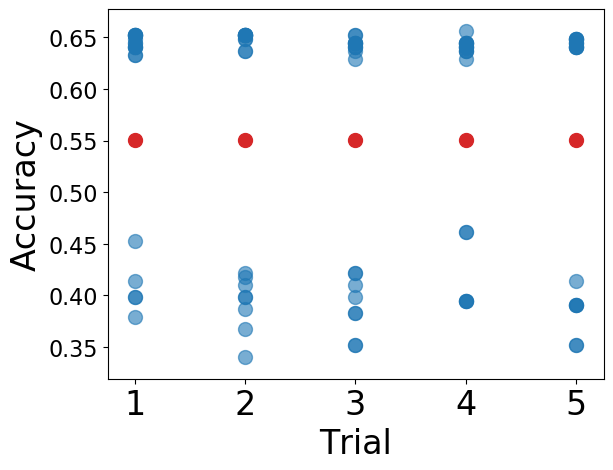}
            \caption{Pythia-1.4B}\label{fig:removal-image5}
        \end{subfigure}
        \hfill
            \begin{subfigure}{0.24\linewidth}
            \centering
            \includegraphics[width=\textwidth]{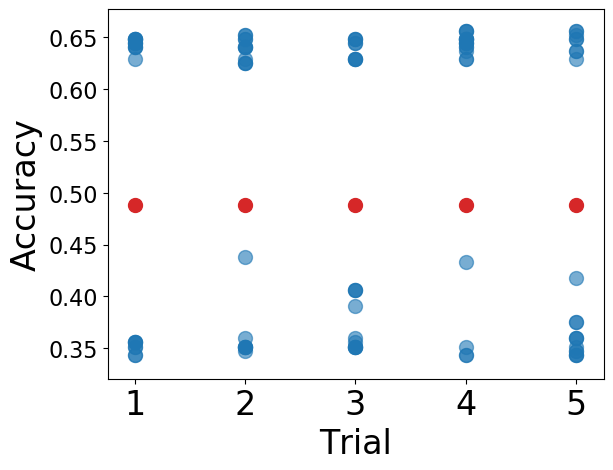}
            \caption{GPT-Neo-2.7B}\label{fig:removal-image6}
        \end{subfigure}
        \hfill
        \begin{subfigure}{0.24\linewidth}
            \centering
            \includegraphics[width=\textwidth]{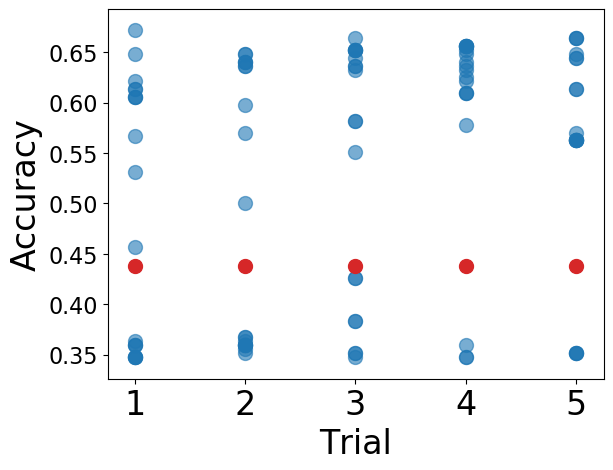}
            \caption{Llama2-7B}\label{fig:removal-image7}
        \end{subfigure}%
        \hfill
        \begin{subfigure}{0.24\linewidth}
            \centering
            \includegraphics[width=\textwidth]{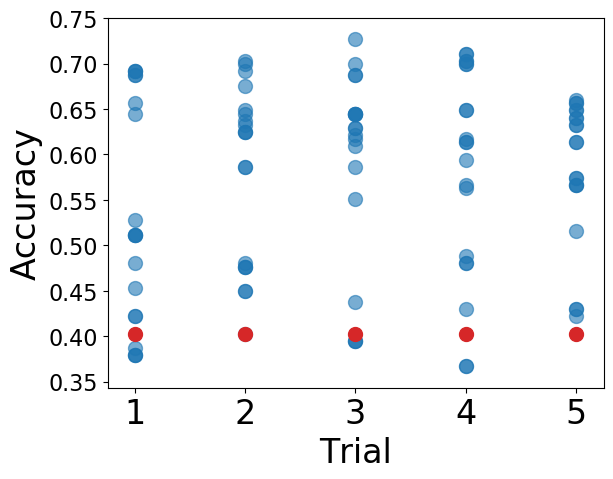}
            \caption{Llama2-13B}\label{fig:removal-image8}
        \end{subfigure}
        \hfill
            \begin{minipage}{.1cm}
            \vfill
            \end{minipage}
    \end{minipage}%
    \hfill
    \begin{minipage}[c]{\linewidth}
        \caption{\label{fig:}One-shot in-context learning performance on the QQP dataset across 5 trials.}
    \end{minipage}
\end{figure*}
\begin{figure*}[h]
    \centering
    \begin{minipage}[t]{\linewidth}
        \begin{subfigure}{0.24\linewidth}
            \centering
            \includegraphics[width=\textwidth]{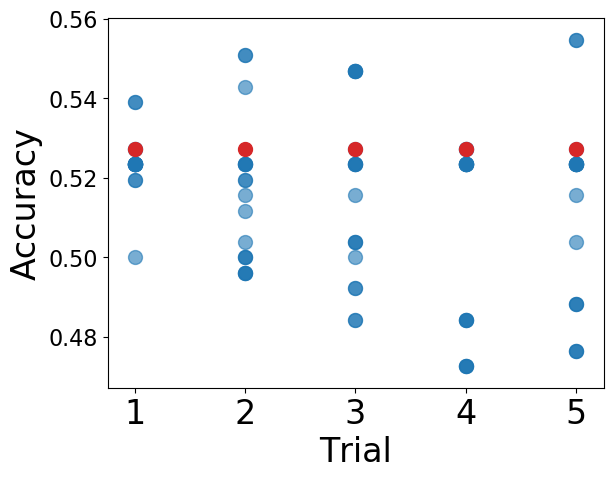}
            \caption{Pythia-160M}\label{fig:removal-image1}
        \end{subfigure}%
        \hfill
        \begin{subfigure}{0.24\linewidth}
            \centering
            \includegraphics[width=\textwidth]{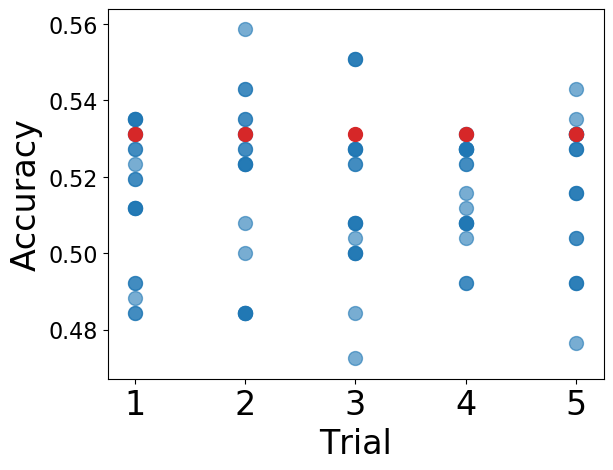}
            \caption{Opt-350M}\label{fig:removal-image2}
        \end{subfigure}
        \hfill
        \begin{subfigure}{0.24\linewidth}
            \centering
            \includegraphics[width=\textwidth]{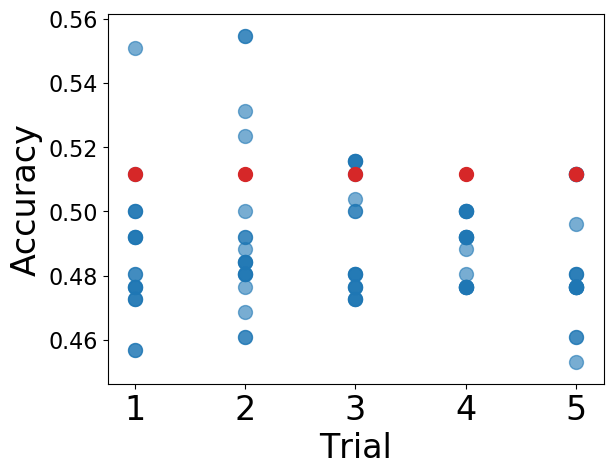}
            \caption{Opt-1.3B}\label{fig:removal-image3}
        \end{subfigure}
        \begin{subfigure}{0.24\linewidth}
            \centering
            \includegraphics[width=\textwidth]{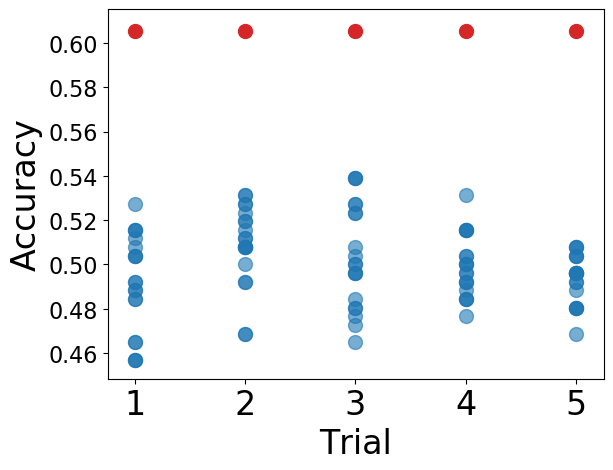}
            \caption{GPT-Neo-1.3B}\label{fig:removal-image4}
        \end{subfigure}%
        \hfill
        \begin{subfigure}{0.24\linewidth}
            \centering
            \includegraphics[width=\textwidth]{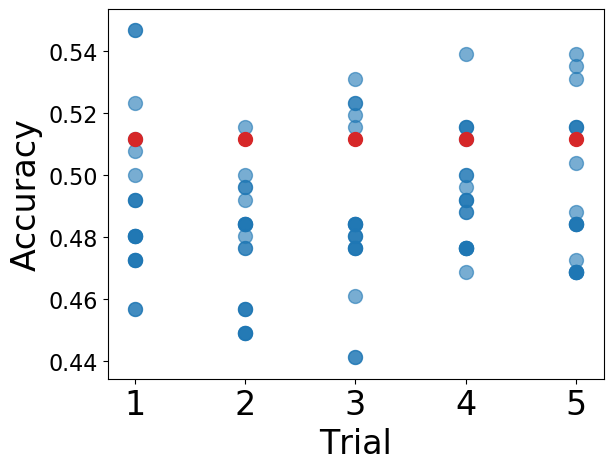}
            \caption{Pythia-1.4B}\label{fig:removal-image5}
        \end{subfigure}
        \hfill
            \begin{subfigure}{0.24\linewidth}
            \centering
            \includegraphics[width=\textwidth]{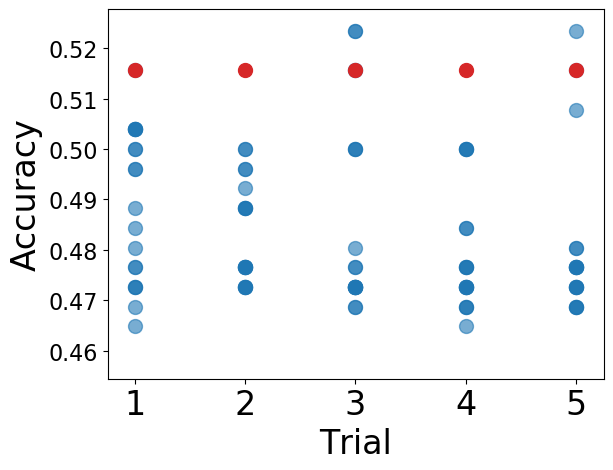}
            \caption{GPT-Neo-2.7B}\label{fig:removal-image6}
        \end{subfigure}
        \hfill
        \begin{subfigure}{0.24\linewidth}
            \centering
            \includegraphics[width=\textwidth]{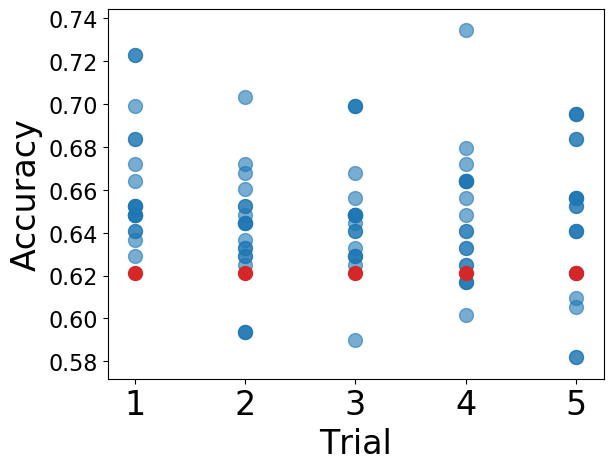}
            \caption{Llama2-7B}\label{fig:removal-image7}
        \end{subfigure}%
        \hfill
        \begin{subfigure}{0.24\linewidth}
            \centering
            \includegraphics[width=\textwidth]{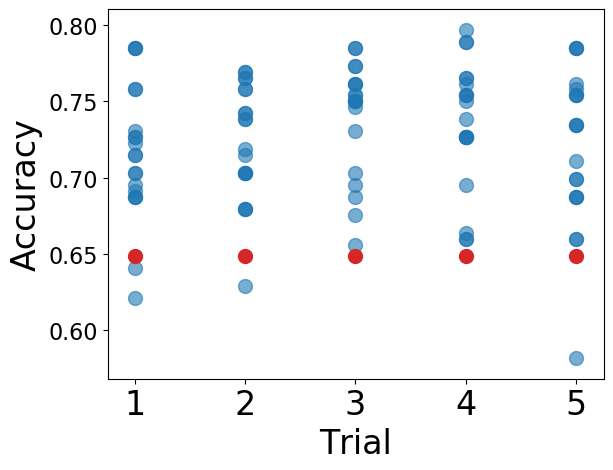}
            \caption{Llama2-13B}\label{fig:removal-image8}
        \end{subfigure}
        \hfill
            \begin{minipage}{.1cm}
            \vfill
            \end{minipage}
    \end{minipage}%
    \hfill
    \begin{minipage}[c]{\linewidth}
        \caption{\label{fig:}One-shot in-context learning performance on the RTE dataset across 5 trials.}
    \end{minipage}
\end{figure*}
\begin{figure*}[h]
    \centering
    \begin{minipage}[t]{\linewidth}
        \begin{subfigure}{0.24\linewidth}
            \centering
            \includegraphics[width=\textwidth]{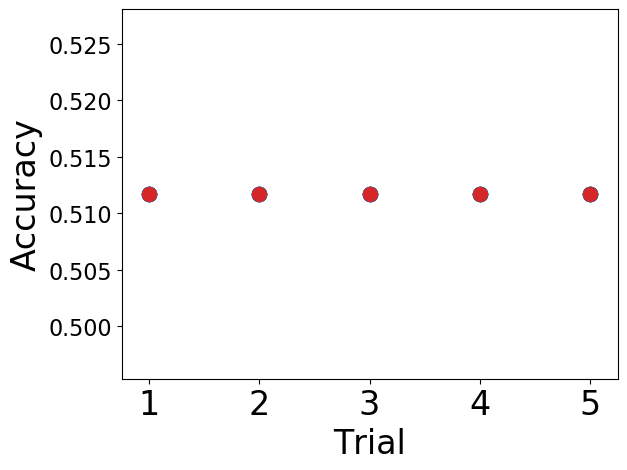}
            \caption{Pythia-160M}\label{fig:removal-image1}
        \end{subfigure}%
        \hfill
        \begin{subfigure}{0.24\linewidth}
            \centering
            \includegraphics[width=\textwidth]{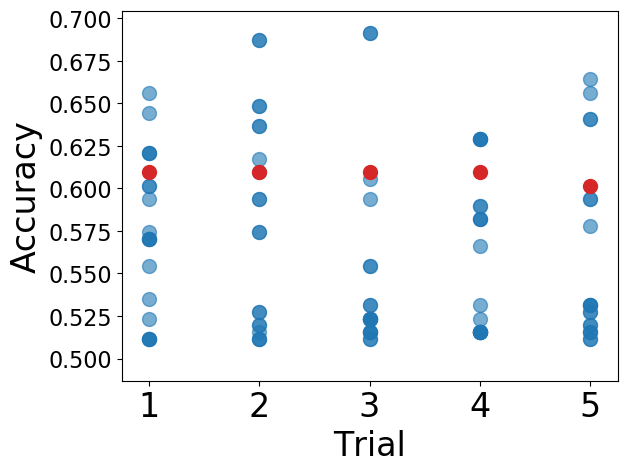}
            \caption{Opt-350M}\label{fig:removal-image2}
        \end{subfigure}
        \hfill
        \begin{subfigure}{0.24\linewidth}
            \centering
            \includegraphics[width=\textwidth]{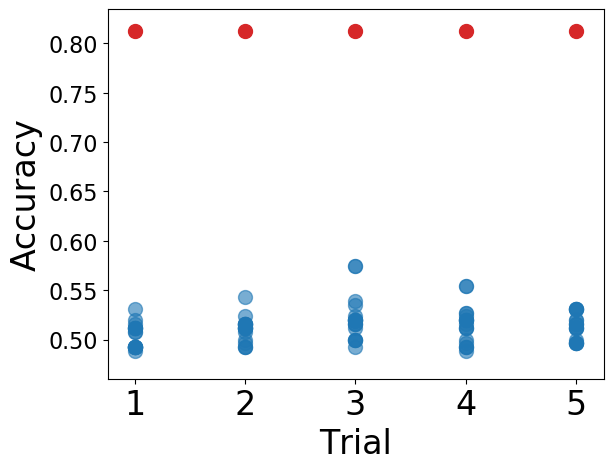}
            \caption{Opt-1.3B}\label{fig:removal-image3}
        \end{subfigure}
        \begin{subfigure}{0.24\linewidth}
            \centering
            \includegraphics[width=\textwidth]{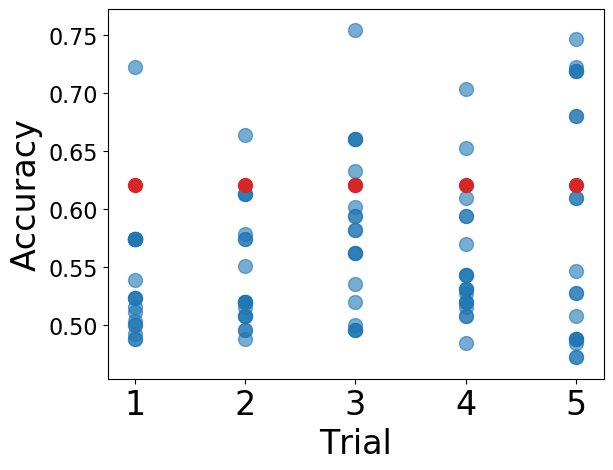}
            \caption{GPT-Neo-1.3B}\label{fig:removal-image4}
        \end{subfigure}%
        \hfill
        \begin{subfigure}{0.24\linewidth}
            \centering
            \includegraphics[width=\textwidth]{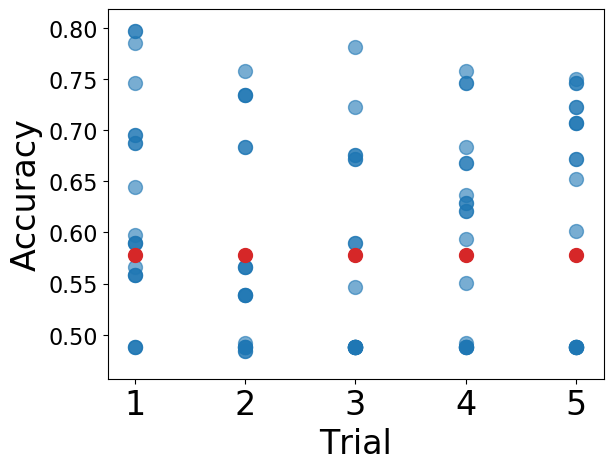}
            \caption{Pythia-1.4B}\label{fig:removal-image5}
        \end{subfigure}
        \hfill
            \begin{subfigure}{0.24\linewidth}
            \centering
            \includegraphics[width=\textwidth]{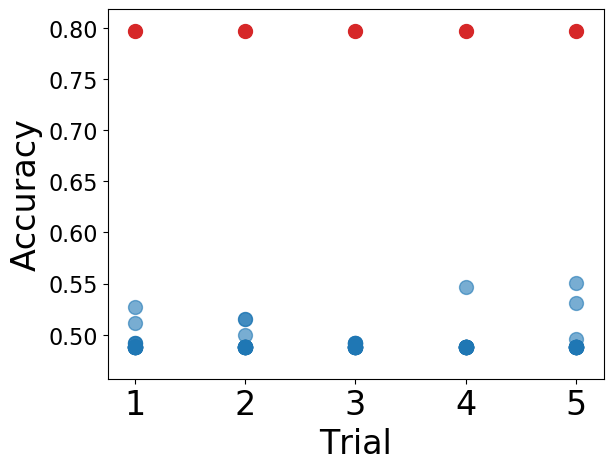}
            \caption{GPT-Neo-2.7B}\label{fig:removal-image6}
        \end{subfigure}
        \hfill
        \begin{subfigure}{0.24\linewidth}
            \centering
            \includegraphics[width=\textwidth]{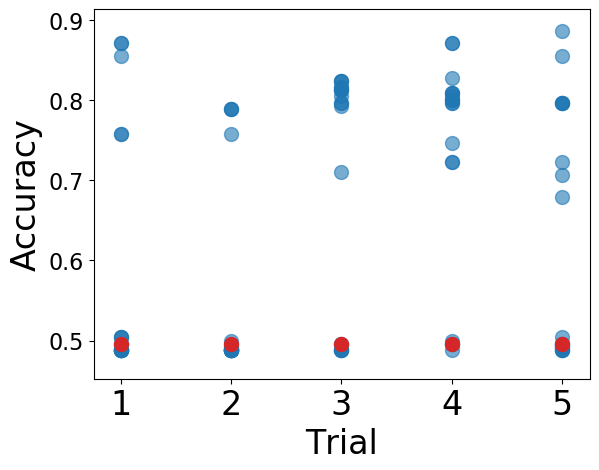}
            \caption{Llama2-7B}\label{fig:removal-image7}
        \end{subfigure}%
        \hfill
        \begin{subfigure}{0.24\linewidth}
            \centering
            \includegraphics[width=\textwidth]{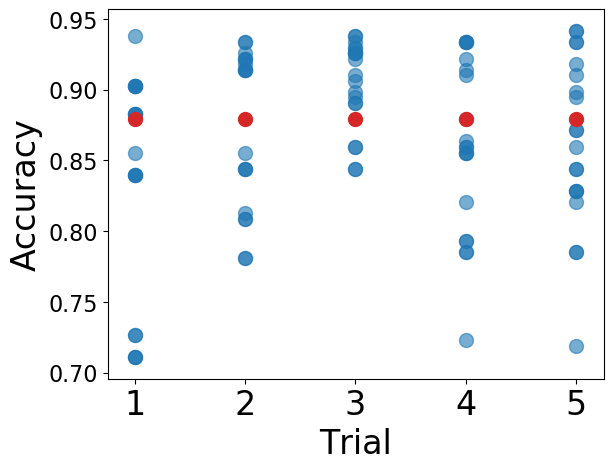}
            \caption{Llama2-13B}\label{fig:removal-image8}
        \end{subfigure}
        \hfill
            \begin{minipage}{.1cm}
            \vfill
            \end{minipage}
    \end{minipage}%
    \hfill
    \begin{minipage}[c]{\linewidth}
        \caption{\label{fig:}One-shot in-context learning performance on the SST-2 dataset across 5 trials.}
    \end{minipage}
\end{figure*}
\begin{figure*}[h]
    \centering
    \begin{minipage}[t]{\linewidth}
        \begin{subfigure}{0.24\linewidth}
            \centering
            \includegraphics[width=\textwidth]{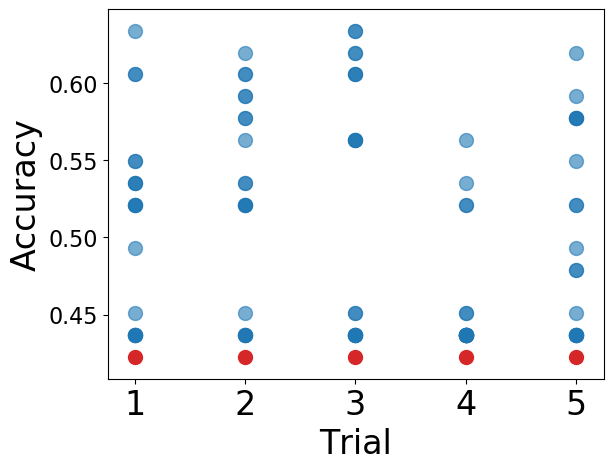}
            \caption{Pythia-160M}\label{fig:removal-image1}
        \end{subfigure}%
        \hfill
        \begin{subfigure}{0.24\linewidth}
            \centering
            \includegraphics[width=\textwidth]{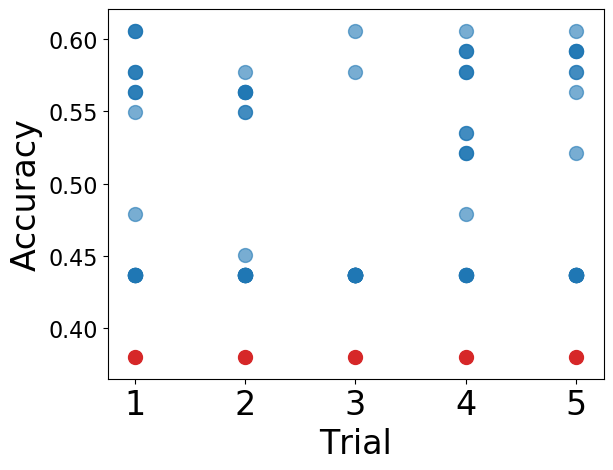}
            \caption{Opt-350M}\label{fig:removal-image2}
        \end{subfigure}
        \hfill
        \begin{subfigure}{0.24\linewidth}
            \centering
            \includegraphics[width=\textwidth]{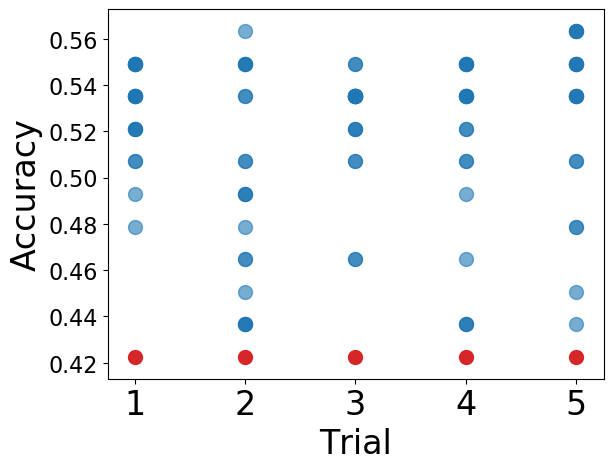}
            \caption{Opt-1.3B}\label{fig:removal-image3}
        \end{subfigure}
        \begin{subfigure}{0.24\linewidth}
            \centering
            \includegraphics[width=\textwidth]{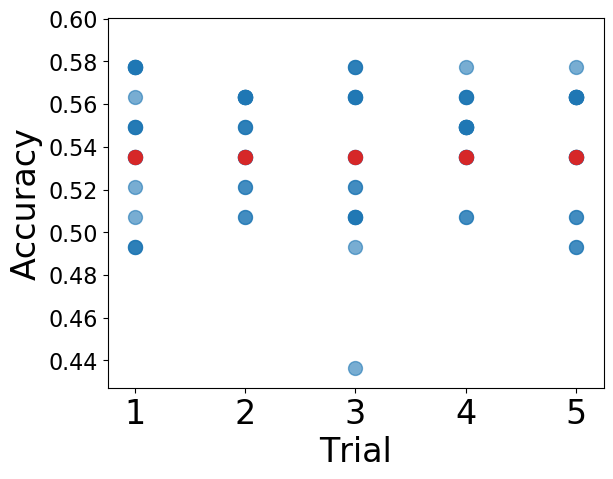}
            \caption{GPT-Neo-1.3B}\label{fig:removal-image4}
        \end{subfigure}%
        \hfill
        \begin{subfigure}{0.24\linewidth}
            \centering
            \includegraphics[width=\textwidth]{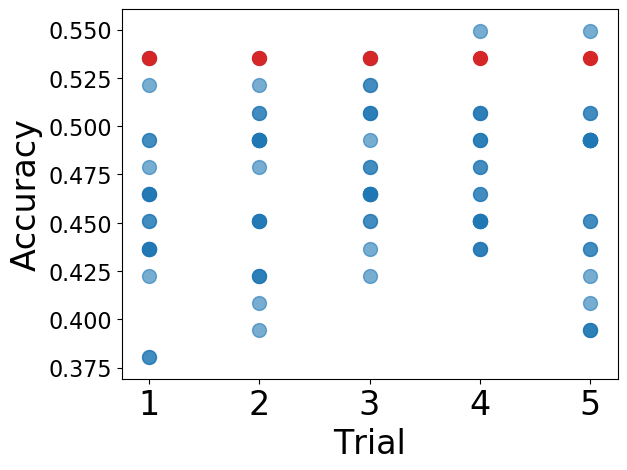}
            \caption{Pythia-1.4B}\label{fig:removal-image5}
        \end{subfigure}
        \hfill
            \begin{subfigure}{0.24\linewidth}
            \centering
            \includegraphics[width=\textwidth]{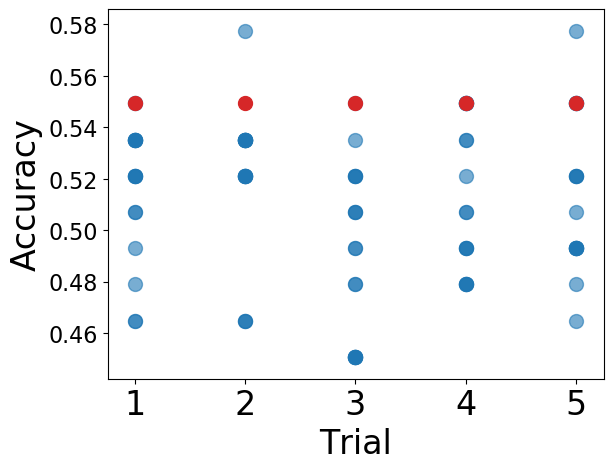}
            \caption{GPT-Neo-2.7B}\label{fig:removal-image6}
        \end{subfigure}
        \hfill
        \begin{subfigure}{0.24\linewidth}
            \centering
            \includegraphics[width=\textwidth]{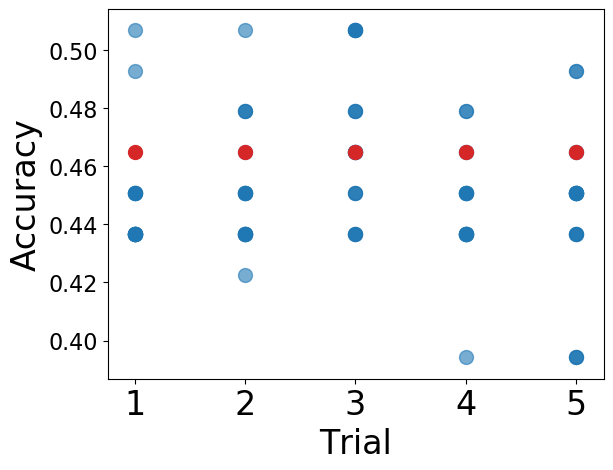}
            \caption{Llama2-7B}\label{fig:removal-image7}
        \end{subfigure}%
        \hfill
        \begin{subfigure}{0.24\linewidth}
            \centering
            \includegraphics[width=\textwidth]{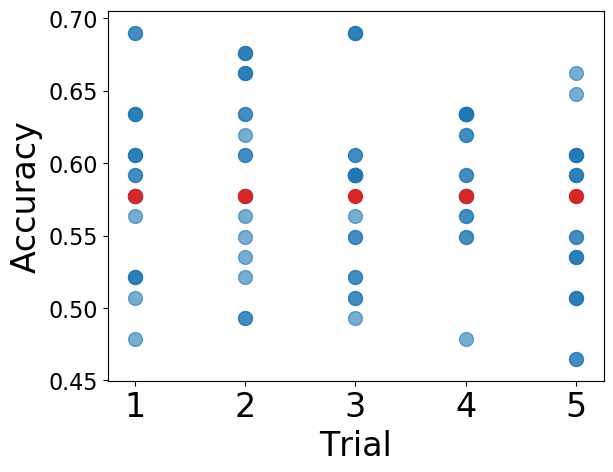}
            \caption{Llama2-13B}\label{fig:removal-image8}
        \end{subfigure}
        \hfill
            \begin{minipage}{.1cm}
            \vfill
            \end{minipage}
    \end{minipage}%
    \hfill
    \begin{minipage}[c]{\linewidth}
        \caption{\label{fig:}One-shot in-context learning performance on the WNLI dataset across 5 trials.}
    \end{minipage}
\end{figure*}

\begin{figure*}[h]
    \centering
    \begin{minipage}[t]{\linewidth}
        \begin{subfigure}{0.24\linewidth}
            \centering
            \includegraphics[width=\textwidth]{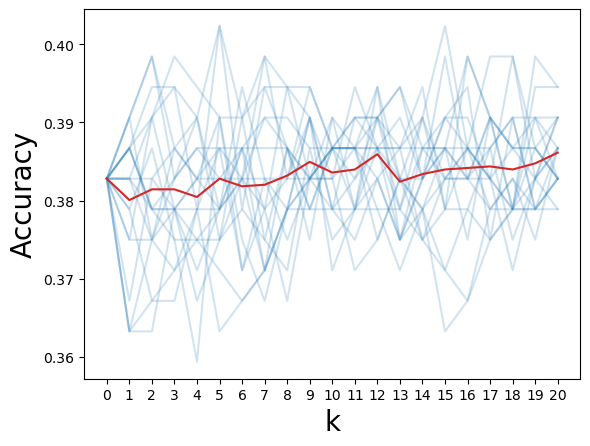}
            \caption{Pythia-160M}\label{fig:removal-image1}
        \end{subfigure}%
        \hfill
        \begin{subfigure}{0.24\linewidth}
            \centering
            \includegraphics[width=\textwidth]{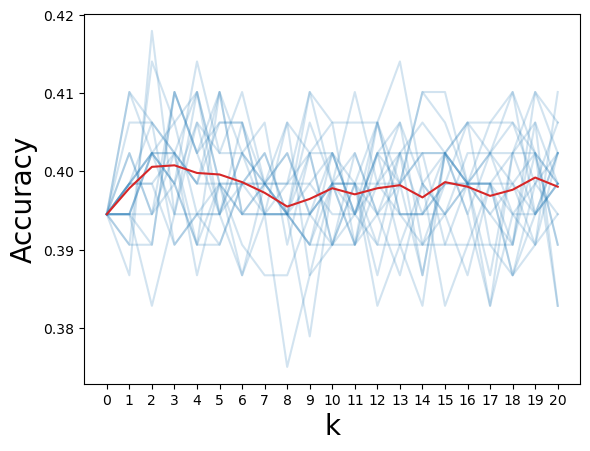}
            \caption{Opt-350M}\label{fig:removal-image2}
        \end{subfigure}
        \hfill
        \begin{subfigure}{0.24\linewidth}
            \centering
            \includegraphics[width=\textwidth]{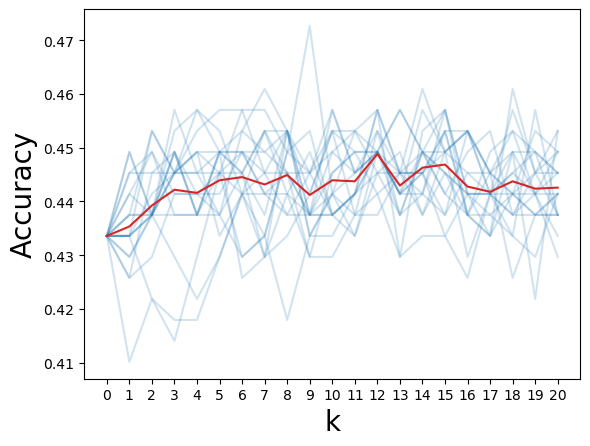}
            \caption{Opt-1.3B}\label{fig:removal-image3}
        \end{subfigure}
        \begin{subfigure}{0.24\linewidth}
            \centering
            \includegraphics[width=\textwidth]{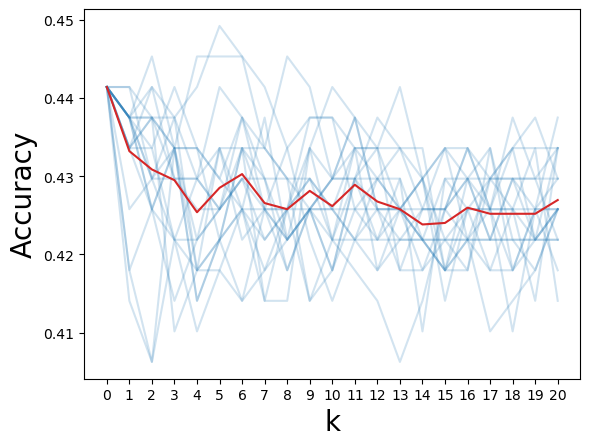}
            \caption{GPT-Neo-1.3B}\label{fig:removal-image4}
        \end{subfigure}%
        \hfill
        \begin{subfigure}{0.24\linewidth}
            \centering
            \includegraphics[width=\textwidth]{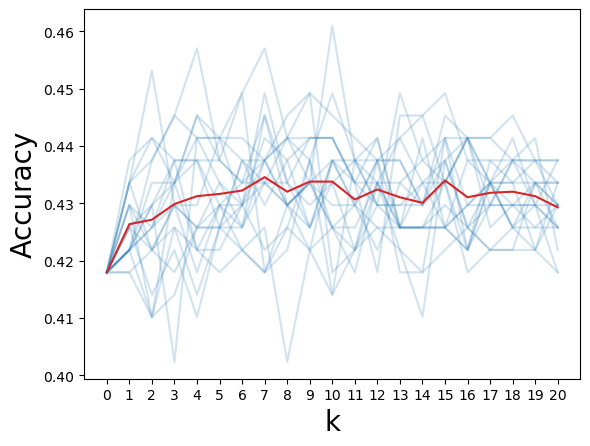}
            \caption{Pythia-1.4B}\label{fig:removal-image5}
        \end{subfigure}
        \hfill
            \begin{subfigure}{0.24\linewidth}
            \centering
            \includegraphics[width=\textwidth]{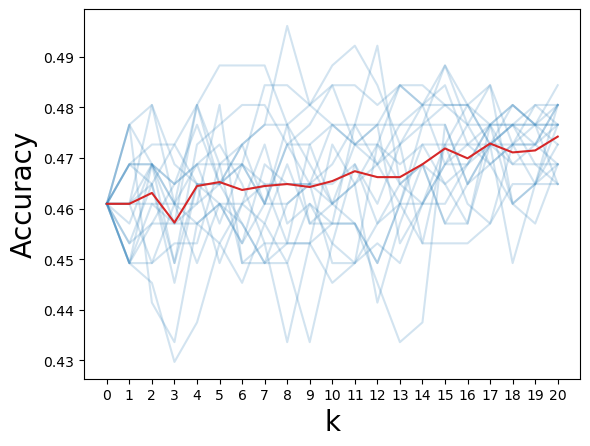}
            \caption{GPT-Neo-2.7B}\label{fig:removal-image6}
        \end{subfigure}
        \hfill
        \begin{subfigure}{0.24\linewidth}
            \centering
            \includegraphics[width=\textwidth]{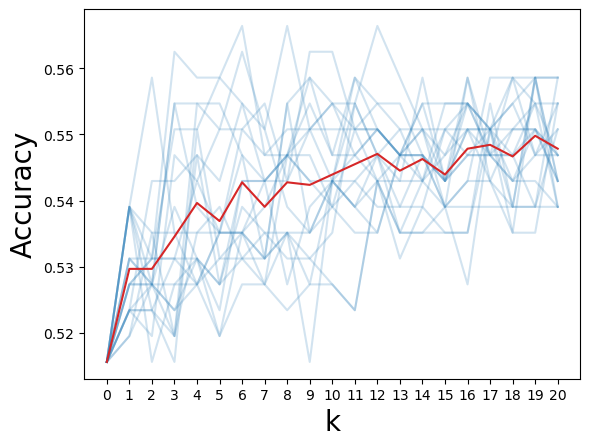}
            \caption{Llama2-7B}\label{fig:removal-image7}
        \end{subfigure}%
        \hfill
        \begin{subfigure}{0.24\linewidth}
            \centering
            \includegraphics[width=\textwidth]{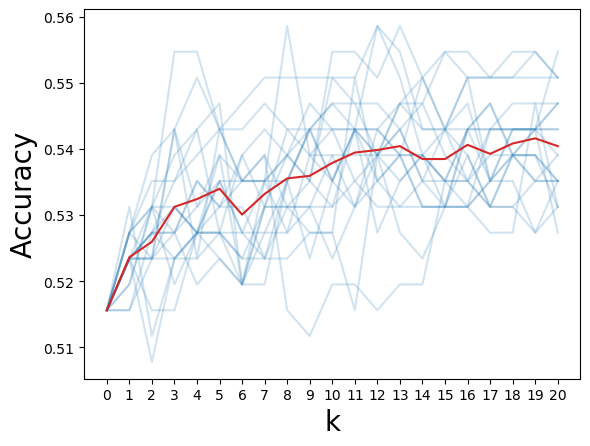}
            \caption{Llama2-13B}\label{fig:removal-image8}
        \end{subfigure}
        \hfill
            \begin{minipage}{.1cm}
            \vfill
            \end{minipage}
    \end{minipage}%
    \hfill
    \begin{minipage}[c]{\linewidth}
        \caption{\label{fig:}Performance of each model on Hellaswag dataset. In each plot, the red line indicates the averages of all permutations for one trial, overlaying blue lines for individual permutations.}
    \end{minipage}
\end{figure*}
\begin{figure*}[h]
    \centering
    \begin{minipage}[t]{\linewidth}
        \begin{subfigure}{0.24\linewidth}
            \centering
            \includegraphics[width=\textwidth]{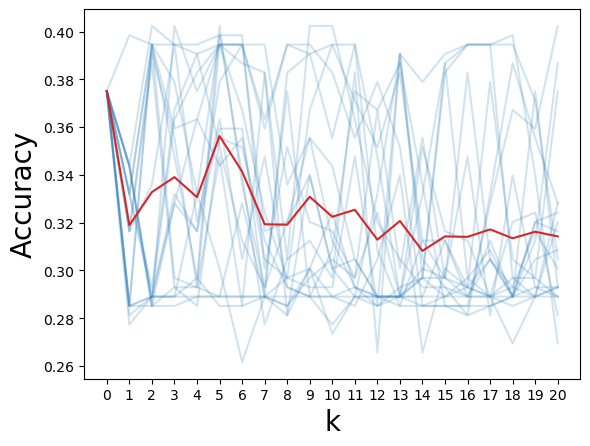}
            \caption{Pythia-160M}\label{fig:removal-image1}
        \end{subfigure}%
        \hfill
        \begin{subfigure}{0.24\linewidth}
            \centering
            \includegraphics[width=\textwidth]{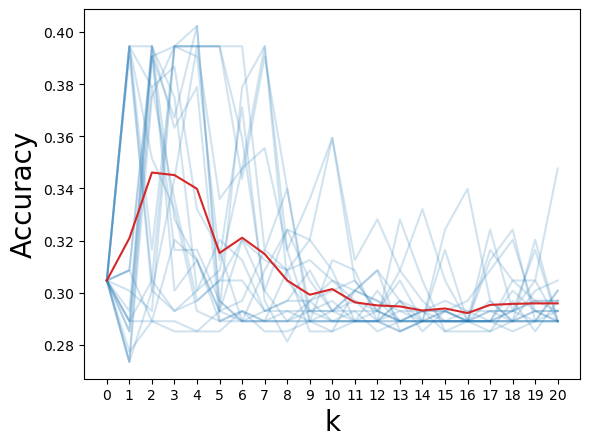}
            \caption{Opt-350M}\label{fig:removal-image2}
        \end{subfigure}
        \hfill
        \begin{subfigure}{0.24\linewidth}
            \centering
            \includegraphics[width=\textwidth]{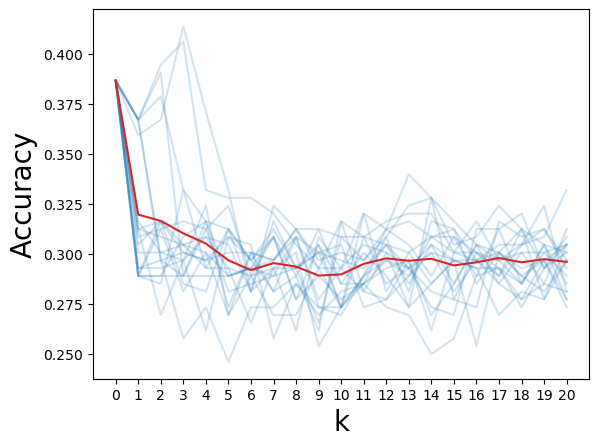}
            \caption{Opt-1.3B}\label{fig:removal-image3}
        \end{subfigure}
        \begin{subfigure}{0.24\linewidth}
            \centering
            \includegraphics[width=\textwidth]{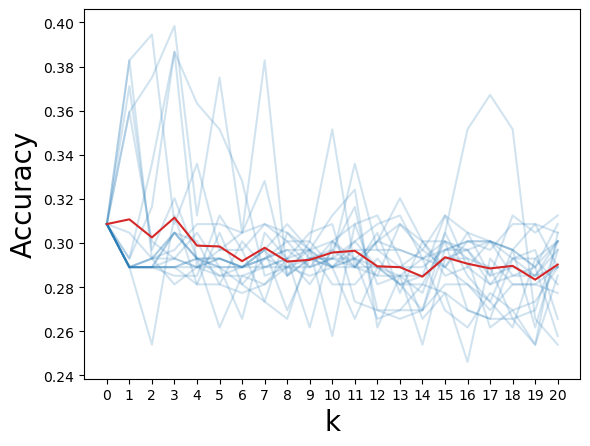}
            \caption{GPT-Neo-1.3B}\label{fig:removal-image4}
        \end{subfigure}%
        \hfill
        \begin{subfigure}{0.24\linewidth}
            \centering
            \includegraphics[width=\textwidth]{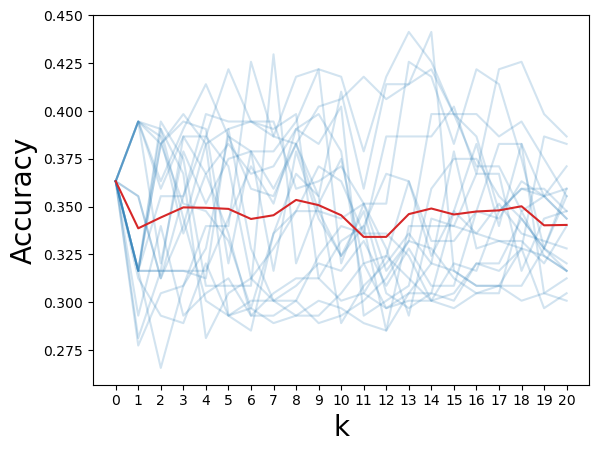}
            \caption{Pythia-1.4B}\label{fig:removal-image5}
        \end{subfigure}
        \hfill
            \begin{subfigure}{0.24\linewidth}
            \centering
            \includegraphics[width=\textwidth]{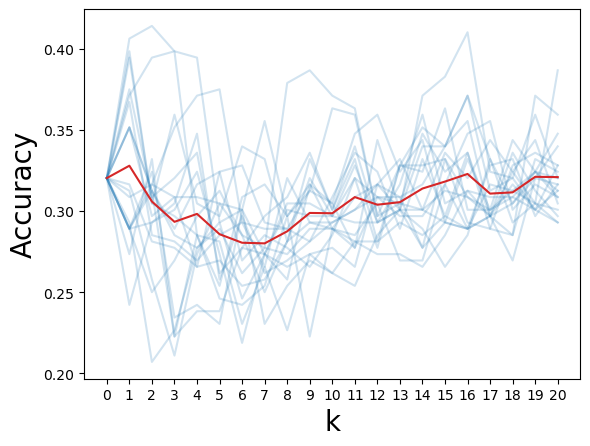}
            \caption{GPT-Neo-2.7B}\label{fig:removal-image6}
        \end{subfigure}
        \hfill
        \begin{subfigure}{0.24\linewidth}
            \centering
            \includegraphics[width=\textwidth]{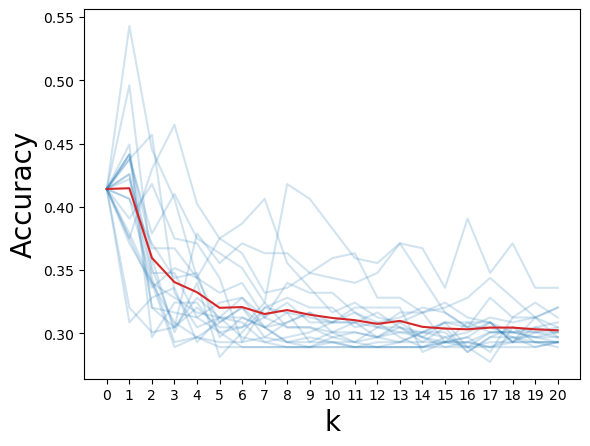}
            \caption{Llama2-7B}\label{fig:removal-image7}
        \end{subfigure}%
        \hfill
        \begin{subfigure}{0.24\linewidth}
            \centering
            \includegraphics[width=\textwidth]{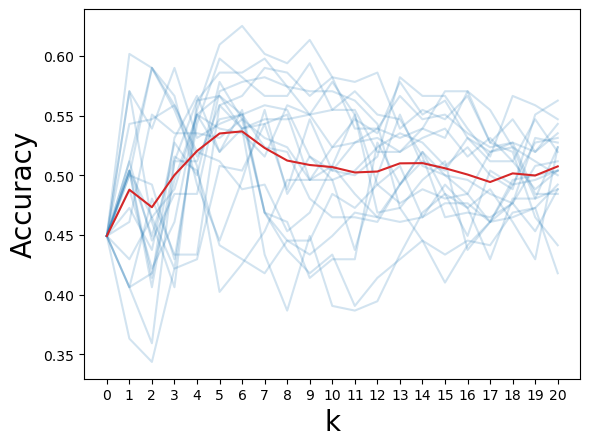}
            \caption{Llama2-13B}\label{fig:removal-image8}
        \end{subfigure}
        \hfill
            \begin{minipage}{.1cm}
            \vfill
            \end{minipage}
    \end{minipage}%
    \hfill
    \begin{minipage}[c]{\linewidth}
        \caption{\label{fig:}Performance of each model on MNLI dataset. In each plot, the red line indicates the averages of all permutations for one trial, overlaying blue lines for individual permutations.}
    \end{minipage}
\end{figure*}
\begin{figure*}[h]
    \centering
    \begin{minipage}[t]{\linewidth}
        \begin{subfigure}{0.24\linewidth}
            \centering
            \includegraphics[width=\textwidth]{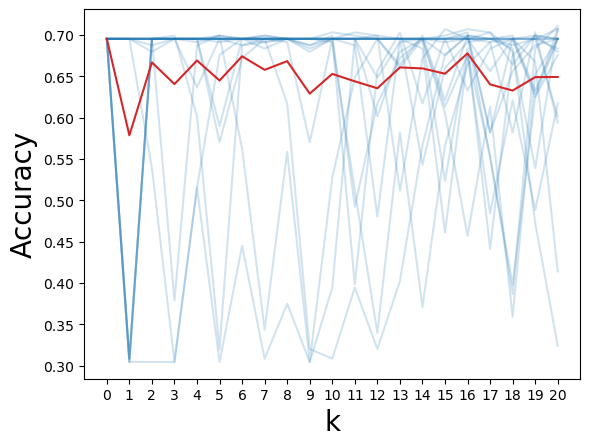}
            \caption{Pythia-160M}\label{fig:removal-image1}
        \end{subfigure}%
        \hfill
        \begin{subfigure}{0.24\linewidth}
            \centering
            \includegraphics[width=\textwidth]{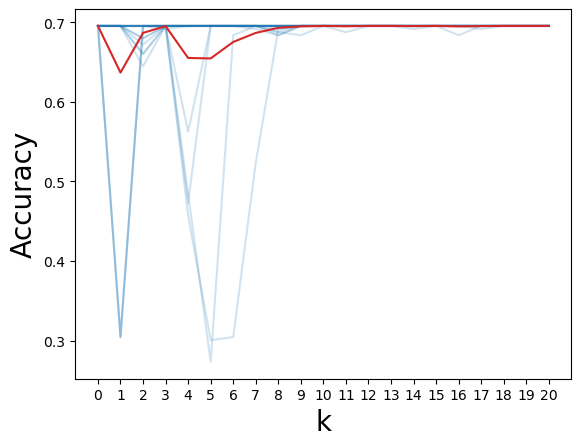}
            \caption{Opt-350M}\label{fig:removal-image2}
        \end{subfigure}
        \hfill
        \begin{subfigure}{0.24\linewidth}
            \centering
            \includegraphics[width=\textwidth]{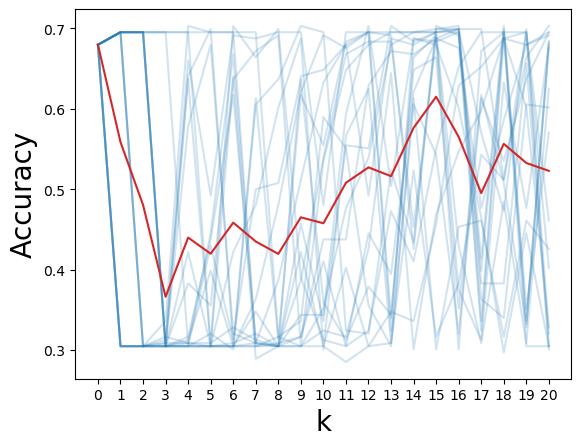}
            \caption{Opt-1.3B}\label{fig:removal-image3}
        \end{subfigure}
        \begin{subfigure}{0.24\linewidth}
            \centering
            \includegraphics[width=\textwidth]{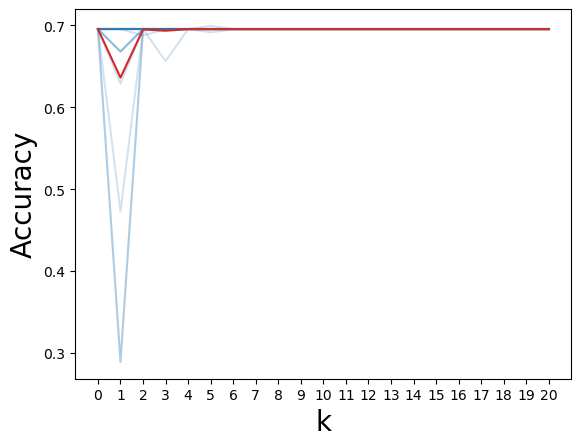}
            \caption{GPT-Neo-1.3B}\label{fig:removal-image4}
        \end{subfigure}%
        \hfill
        \begin{subfigure}{0.24\linewidth}
            \centering
            \includegraphics[width=\textwidth]{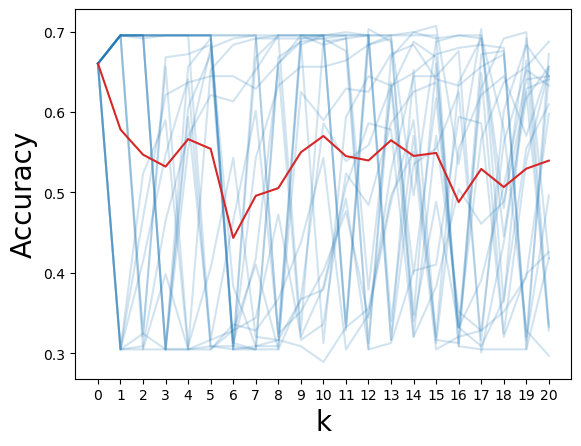}
            \caption{Pythia-1.4B}\label{fig:removal-image5}
        \end{subfigure}
        \hfill
            \begin{subfigure}{0.24\linewidth}
            \centering
            \includegraphics[width=\textwidth]{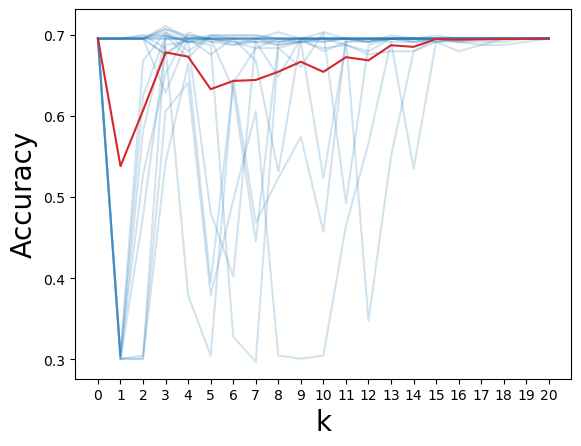}
            \caption{GPT-Neo-2.7B}\label{fig:removal-image6}
        \end{subfigure}
        \hfill
        \begin{subfigure}{0.24\linewidth}
            \centering
            \includegraphics[width=\textwidth]{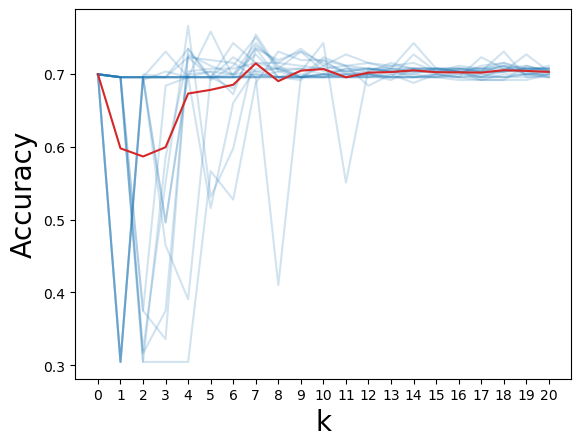}
            \caption{Llama2-7B}\label{fig:removal-image7}
        \end{subfigure}%
        \hfill
        \begin{subfigure}{0.24\linewidth}
            \centering
            \includegraphics[width=\textwidth]{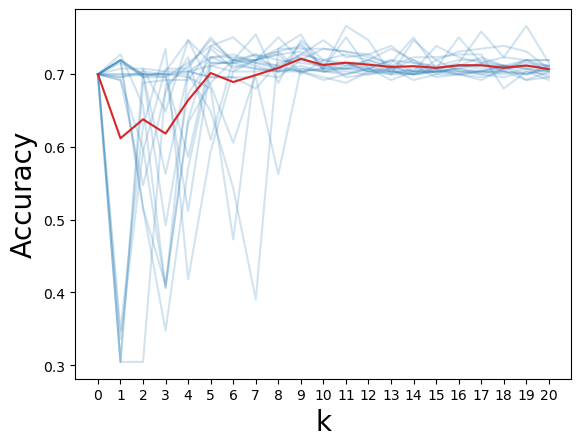}
            \caption{Llama2-13B}\label{fig:removal-image8}
        \end{subfigure}
        \hfill
            \begin{minipage}{.1cm}
            \vfill
            \end{minipage}
    \end{minipage}%
    \hfill
    \begin{minipage}[c]{\linewidth}
        \caption{\label{fig:}Performance of each model on MRPC dataset. In each plot, the red line indicates the averages of all permutations for one trial, overlaying blue lines for individual permutations.}
    \end{minipage}
\end{figure*}
\begin{figure*}[h]
    \centering
    \begin{minipage}[t]{\linewidth}
        \begin{subfigure}{0.24\linewidth}
            \centering
            \includegraphics[width=\textwidth]{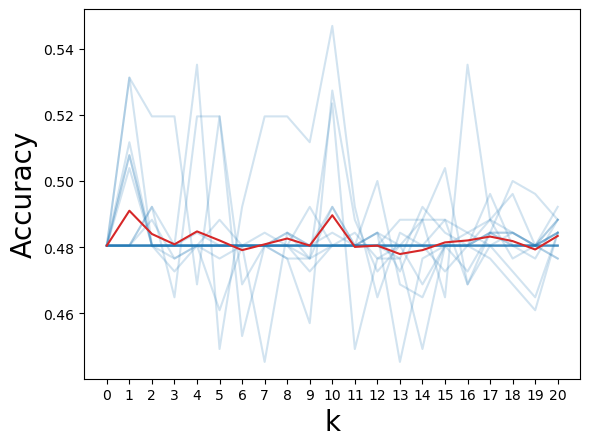}
            \caption{Pythia-160M}\label{fig:removal-image1}
        \end{subfigure}%
        \hfill
        \begin{subfigure}{0.24\linewidth}
            \centering
            \includegraphics[width=\textwidth]{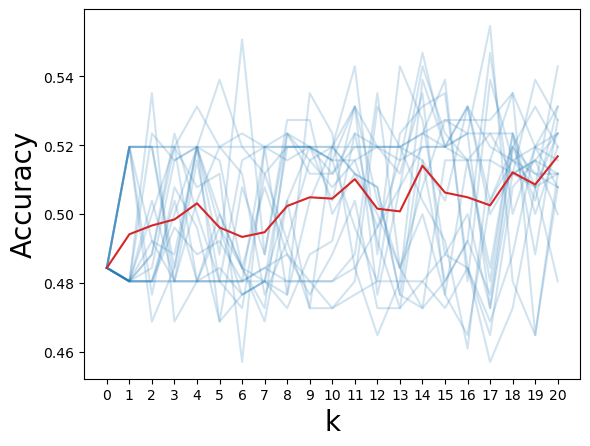}
            \caption{Opt-350M}\label{fig:removal-image2}
        \end{subfigure}
        \hfill
        \begin{subfigure}{0.24\linewidth}
            \centering
            \includegraphics[width=\textwidth]{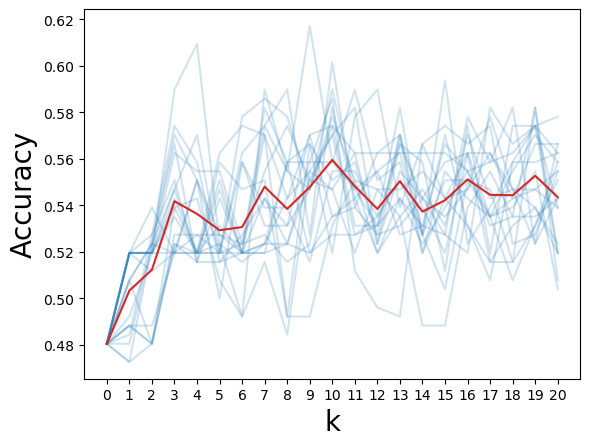}
            \caption{Opt-1.3B}\label{fig:removal-image3}
        \end{subfigure}
        \begin{subfigure}{0.24\linewidth}
            \centering
            \includegraphics[width=\textwidth]{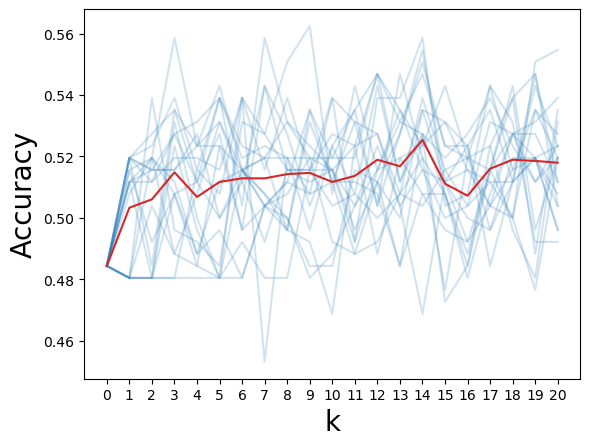}
            \caption{GPT-Neo-1.3B}\label{fig:removal-image4}
        \end{subfigure}%
        \hfill
        \begin{subfigure}{0.24\linewidth}
            \centering
            \includegraphics[width=\textwidth]{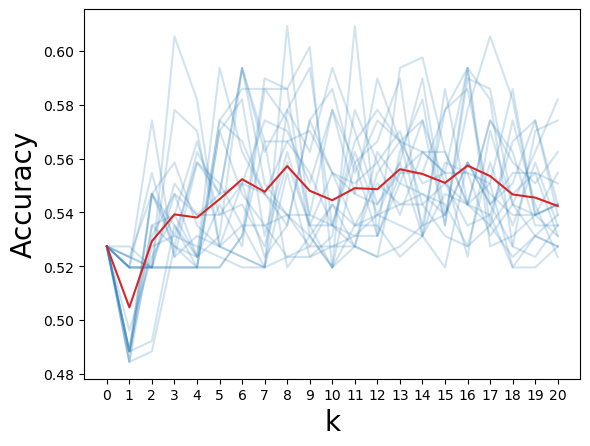}
            \caption{Pythia-1.4B}\label{fig:removal-image5}
        \end{subfigure}
        \hfill
            \begin{subfigure}{0.24\linewidth}
            \centering
            \includegraphics[width=\textwidth]{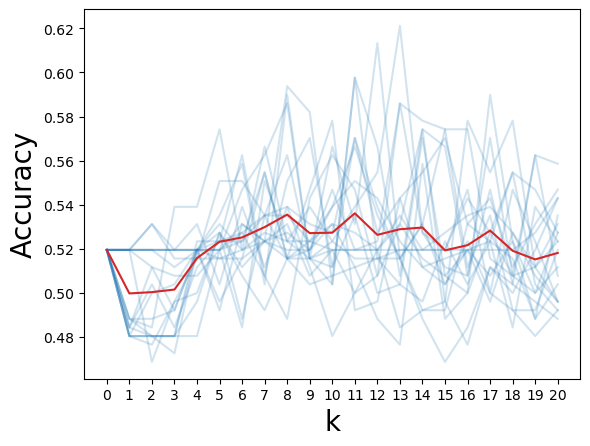}
            \caption{GPT-Neo-2.7B}\label{fig:removal-image6}
        \end{subfigure}
        \hfill
        \begin{subfigure}{0.24\linewidth}
            \centering
            \includegraphics[width=\textwidth]{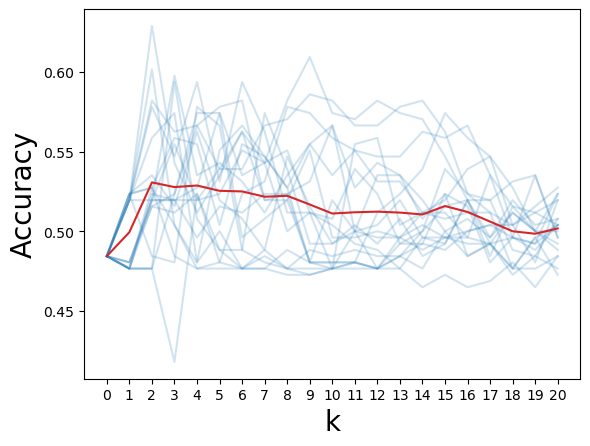}
            \caption{Llama2-7B}\label{fig:removal-image7}
        \end{subfigure}%
        \hfill
        \begin{subfigure}{0.24\linewidth}
            \centering
            \includegraphics[width=\textwidth]{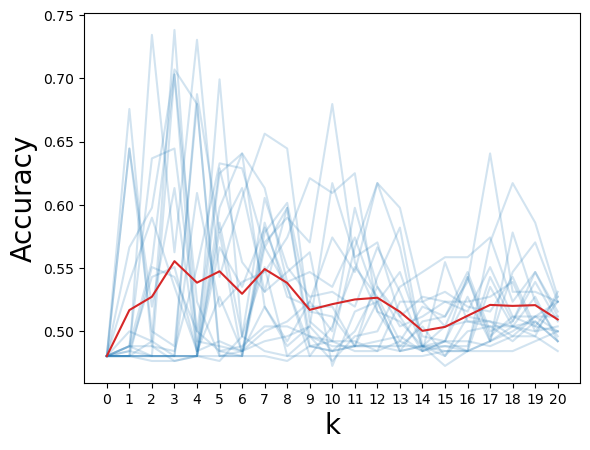}
            \caption{Llama2-13B}\label{fig:removal-image8}
        \end{subfigure}
        \hfill
            \begin{minipage}{.1cm}
            \vfill
            \end{minipage}
    \end{minipage}%
    \hfill
    \begin{minipage}[c]{\linewidth}
        \caption{\label{fig:}Performance of each model on QNLI dataset. In each plot, the red line indicates the averages of all permutations for one trial, overlaying blue lines for individual permutations.}
    \end{minipage}
\end{figure*}
\begin{figure*}[h]
    \centering
    \begin{minipage}[t]{\linewidth}
        \begin{subfigure}{0.24\linewidth}
            \centering
            \includegraphics[width=\textwidth]{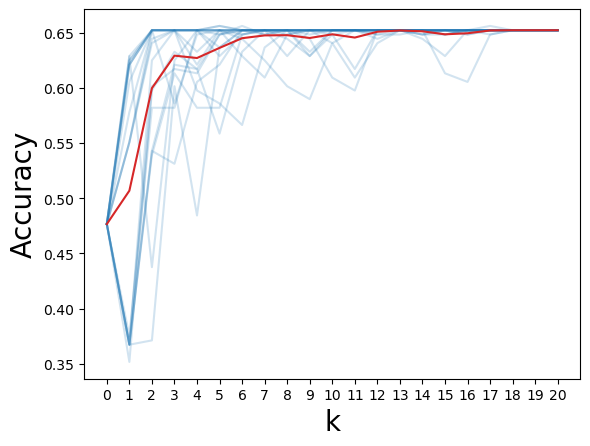}
            \caption{Pythia-160M}\label{fig:removal-image1}
        \end{subfigure}%
        \hfill
        \begin{subfigure}{0.24\linewidth}
            \centering
            \includegraphics[width=\textwidth]{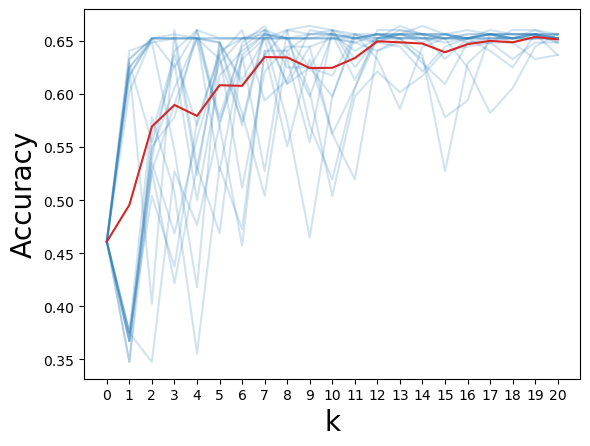}
            \caption{Opt-350M}\label{fig:removal-image2}
        \end{subfigure}
        \hfill
        \begin{subfigure}{0.24\linewidth}
            \centering
            \includegraphics[width=\textwidth]{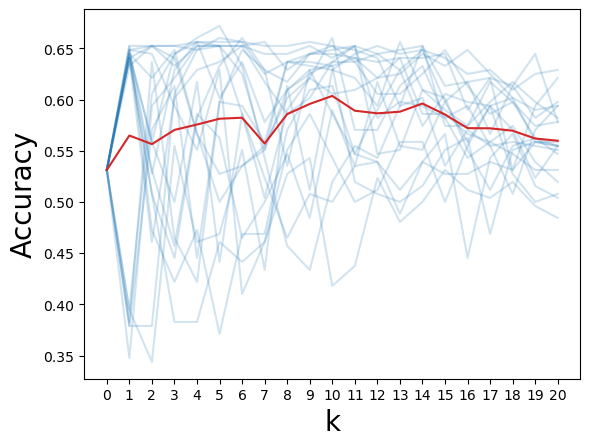}
            \caption{Opt-1.3B}\label{fig:removal-image3}
        \end{subfigure}
        \begin{subfigure}{0.24\linewidth}
            \centering
            \includegraphics[width=\textwidth]{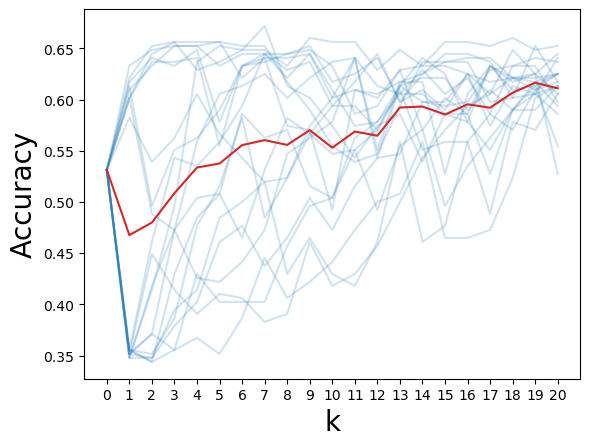}
            \caption{GPT-Neo-1.3B}\label{fig:removal-image4}
        \end{subfigure}%
        \hfill
        \begin{subfigure}{0.24\linewidth}
            \centering
            \includegraphics[width=\textwidth]{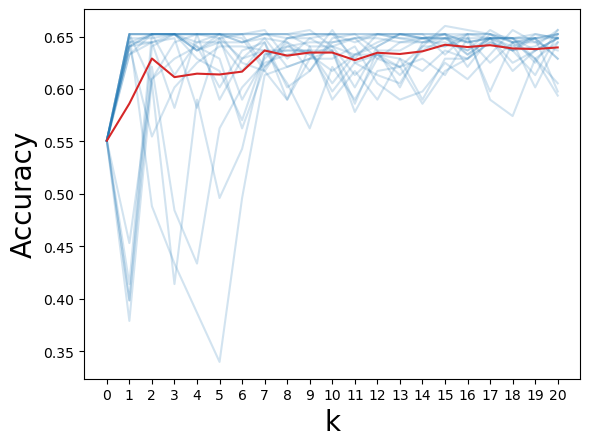}
            \caption{Pythia-1.4B}\label{fig:removal-image5}
        \end{subfigure}
        \hfill
            \begin{subfigure}{0.24\linewidth}
            \centering
            \includegraphics[width=\textwidth]{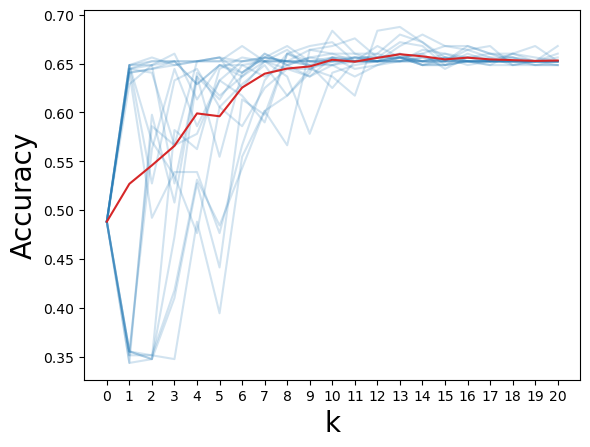}
            \caption{GPT-Neo-2.7B}\label{fig:removal-image6}
        \end{subfigure}
        \hfill
        \begin{subfigure}{0.24\linewidth}
            \centering
            \includegraphics[width=\textwidth]{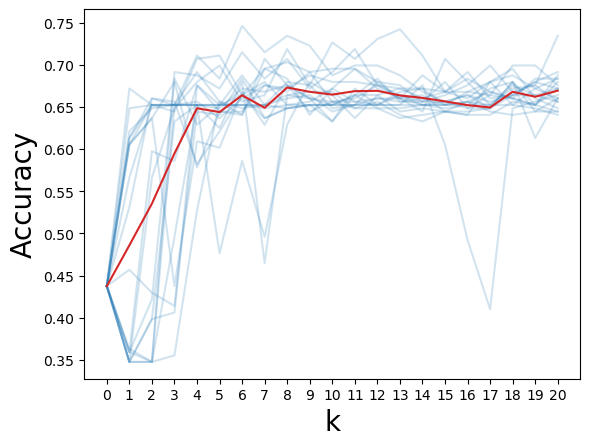}
            \caption{Llama2-7B}\label{fig:removal-image7}
        \end{subfigure}%
        \hfill
        \begin{subfigure}{0.24\linewidth}
            \centering
            \includegraphics[width=\textwidth]{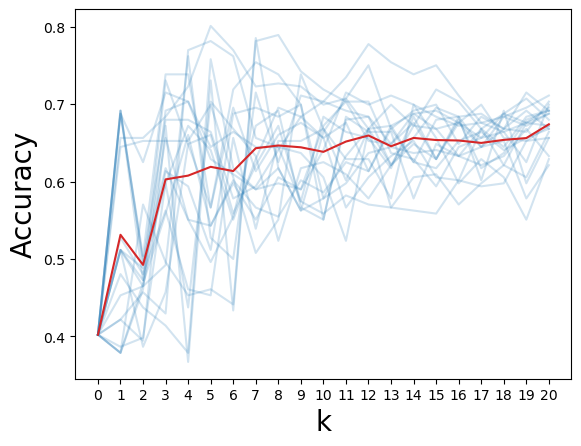}
            \caption{Llama2-13B}\label{fig:removal-image8}
        \end{subfigure}
        \hfill
            \begin{minipage}{.1cm}
            \vfill
            \end{minipage}
    \end{minipage}%
    \hfill
    \begin{minipage}[c]{\linewidth}
        \caption{\label{fig:}Performance of each model on QQP dataset. In each plot, the red line indicates the averages of all permutations for one trial, overlaying blue lines for individual permutations.}
    \end{minipage}
\end{figure*}
\begin{figure*}[h]
    \centering
    \begin{minipage}[t]{\linewidth}
        \begin{subfigure}{0.24\linewidth}
            \centering
            \includegraphics[width=\textwidth]{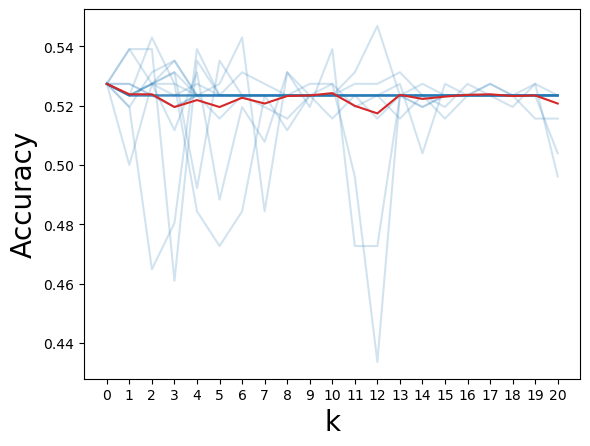}
            \caption{Pythia-160M}\label{fig:removal-image1}
        \end{subfigure}%
        \hfill
        \begin{subfigure}{0.24\linewidth}
            \centering
            \includegraphics[width=\textwidth]{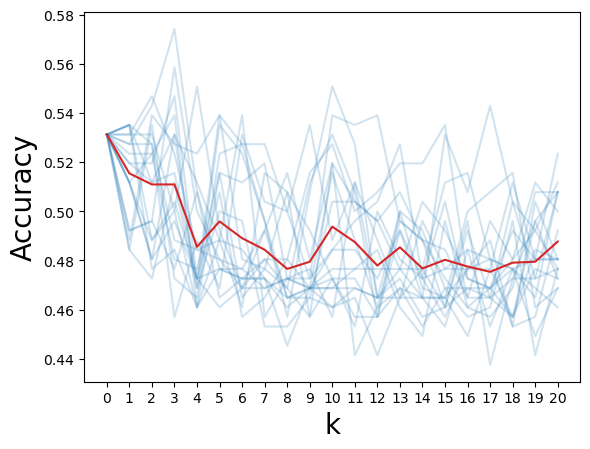}
            \caption{Opt-350M}\label{fig:removal-image2}
        \end{subfigure}
        \hfill
        \begin{subfigure}{0.24\linewidth}
            \centering
            \includegraphics[width=\textwidth]{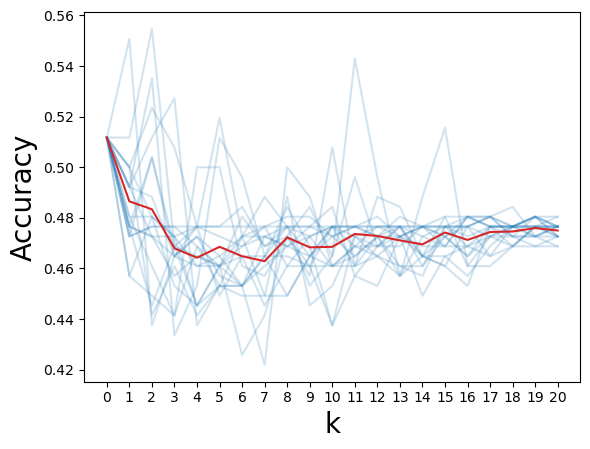}
            \caption{Opt-1.3B}\label{fig:removal-image3}
        \end{subfigure}
        \begin{subfigure}{0.24\linewidth}
            \centering
            \includegraphics[width=\textwidth]{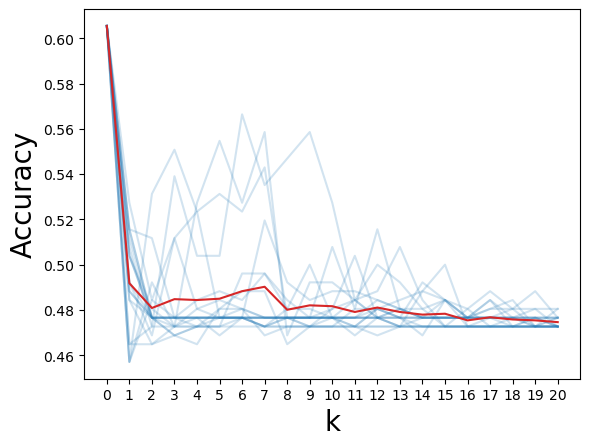}
            \caption{GPT-Neo-1.3B}\label{fig:removal-image4}
        \end{subfigure}%
        \hfill
        \begin{subfigure}{0.24\linewidth}
            \centering
            \includegraphics[width=\textwidth]{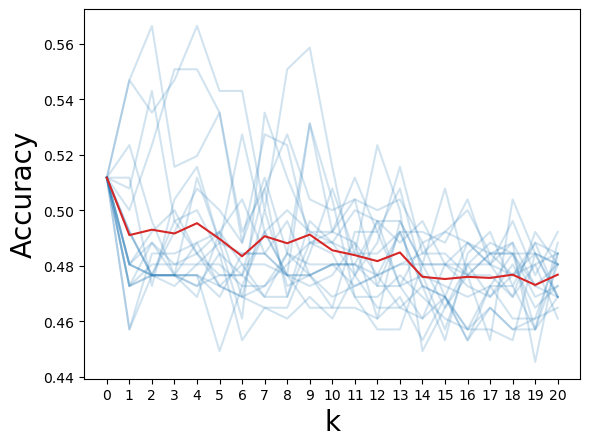}
            \caption{Pythia-1.4B}\label{fig:removal-image5}
        \end{subfigure}
        \hfill
            \begin{subfigure}{0.24\linewidth}
            \centering
            \includegraphics[width=\textwidth]{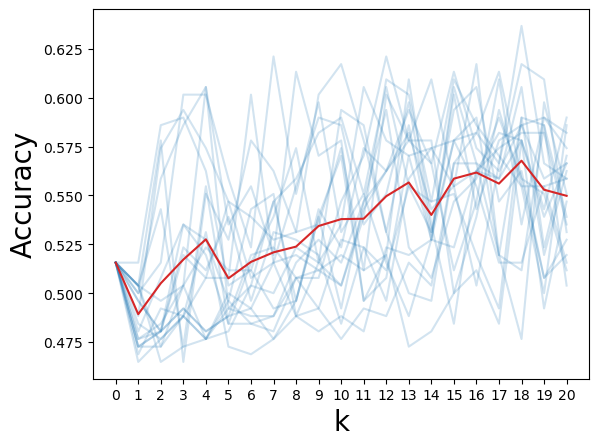}
            \caption{GPT-Neo-2.7B}\label{fig:removal-image6}
        \end{subfigure}
        \hfill
        \begin{subfigure}{0.24\linewidth}
            \centering
            \includegraphics[width=\textwidth]{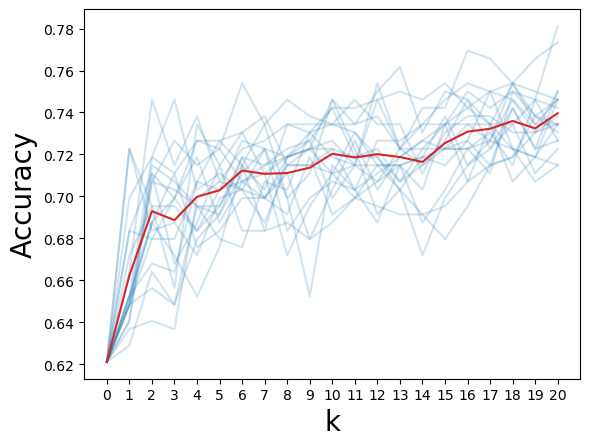}
            \caption{Llama2-7B}\label{fig:removal-image7}
        \end{subfigure}%
        \hfill
        \begin{subfigure}{0.24\linewidth}
            \centering
            \includegraphics[width=\textwidth]{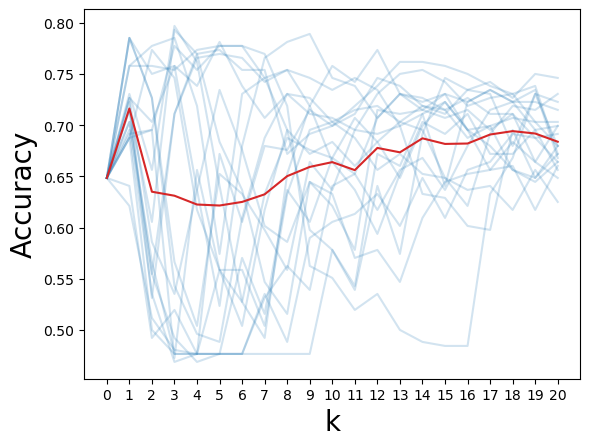}
            \caption{Llama2-13B}\label{fig:removal-image8}
        \end{subfigure}
        \hfill
            \begin{minipage}{.1cm}
            \vfill
            \end{minipage}
    \end{minipage}%
    \hfill
    \begin{minipage}[c]{\linewidth}
        \caption{\label{fig:}Performance of each model on RTE dataset. In each plot, the red line indicates the averages of all permutations for one trial, overlaying blue lines for individual permutations.}
    \end{minipage}
\end{figure*}
\begin{figure*}[h]
    \centering
    \begin{minipage}[t]{\linewidth}
        \begin{subfigure}{0.24\linewidth}
            \centering
            \includegraphics[width=\textwidth]{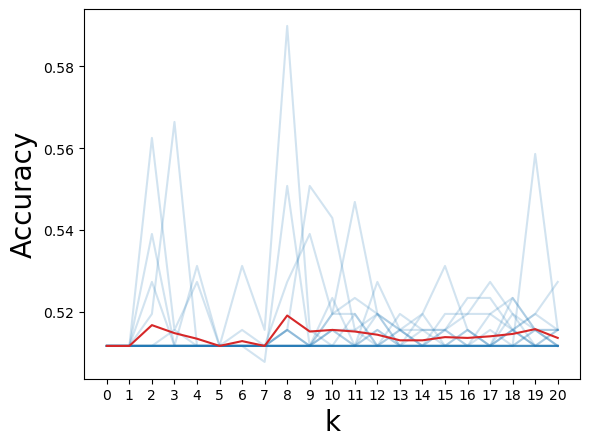}
            \caption{Pythia-160M}\label{fig:removal-image1}
        \end{subfigure}%
        \hfill
        \begin{subfigure}{0.24\linewidth}
            \centering
            \includegraphics[width=\textwidth]{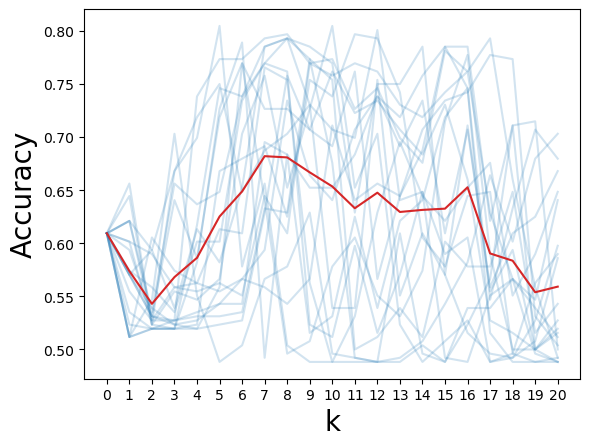}
            \caption{Opt-350M}\label{fig:removal-image2}
        \end{subfigure}
        \hfill
        \begin{subfigure}{0.24\linewidth}
            \centering
            \includegraphics[width=\textwidth]{figures/sst2-opt-1.3b.png}
            \caption{Opt-1.3B}\label{fig:removal-image3}
        \end{subfigure}
        \begin{subfigure}{0.24\linewidth}
            \centering
            \includegraphics[width=\textwidth]{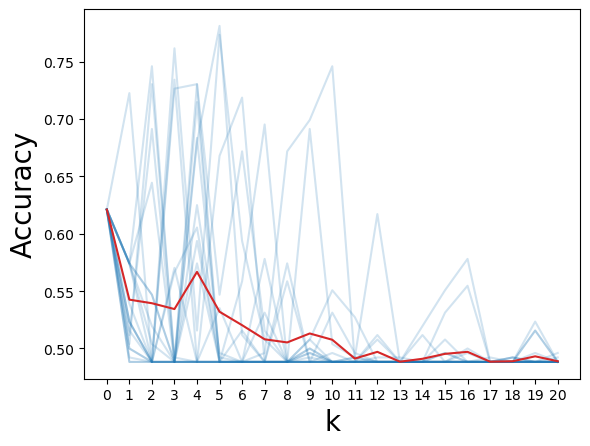}
            \caption{GPT-Neo-1.3B}\label{fig:removal-image4}
        \end{subfigure}%
        \hfill
        \begin{subfigure}{0.24\linewidth}
            \centering
            \includegraphics[width=\textwidth]{figures/sst2-pythia-1.4b.png}
            \caption{Pythia-1.4B}\label{fig:removal-image5}
        \end{subfigure}
        \hfill
            \begin{subfigure}{0.24\linewidth}
            \centering
            \includegraphics[width=\textwidth]{figures/sst2-gpt-neo-2.7b.png}
            \caption{GPT-Neo-2.7B}\label{fig:removal-image6}
        \end{subfigure}
        \hfill
        \begin{subfigure}{0.24\linewidth}
            \centering
            \includegraphics[width=\textwidth]{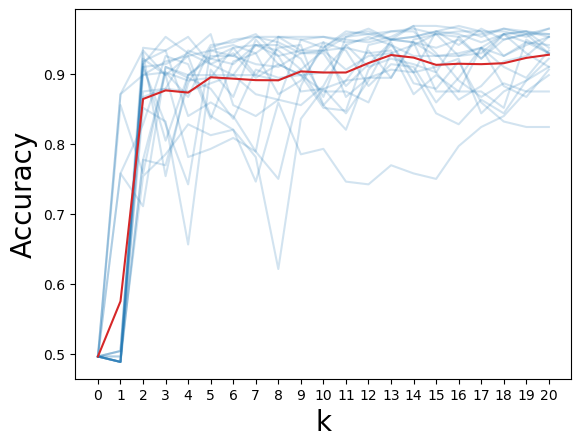}
            \caption{Llama2-7B}\label{fig:removal-image7}
        \end{subfigure}%
        \hfill
        \begin{subfigure}{0.24\linewidth}
            \centering
            \includegraphics[width=\textwidth]{figures/sst2-llama2-13b.png}
            \caption{Llama2-13B}\label{fig:removal-image8}
        \end{subfigure}
        \hfill
            \begin{minipage}{.1cm}
            \vfill
            \end{minipage}
    \end{minipage}%
    \hfill
    \begin{minipage}[c]{\linewidth}
        \caption{\label{fig:}Performance of each model on SST-2 dataset. In each plot, the red line indicates the averages of all permutations for one trial, overlaying blue lines for individual permutations.}
    \end{minipage}
\end{figure*}
\begin{figure*}[h]
    \centering
    \begin{minipage}[t]{\linewidth}
        \begin{subfigure}{0.24\linewidth}
            \centering
            \includegraphics[width=\textwidth]{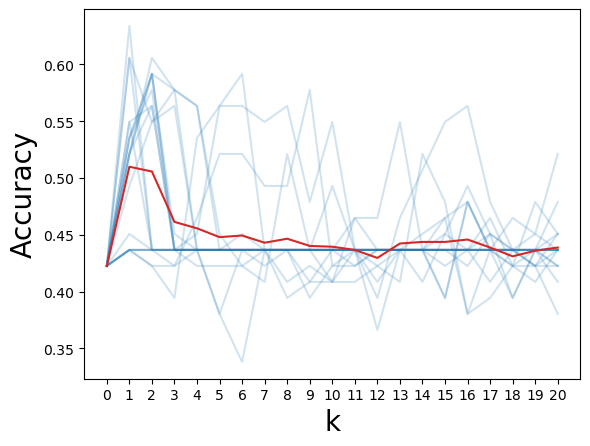}
            \caption{Pythia-160M}\label{fig:removal-image1}
        \end{subfigure}%
        \hfill
        \begin{subfigure}{0.24\linewidth}
            \centering
            \includegraphics[width=\textwidth]{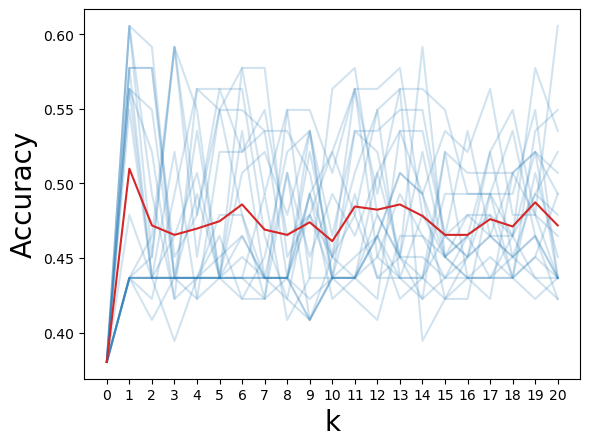}
            \caption{Opt-350M}\label{fig:removal-image2}
        \end{subfigure}
        \hfill
        \begin{subfigure}{0.24\linewidth}
            \centering
            \includegraphics[width=\textwidth]{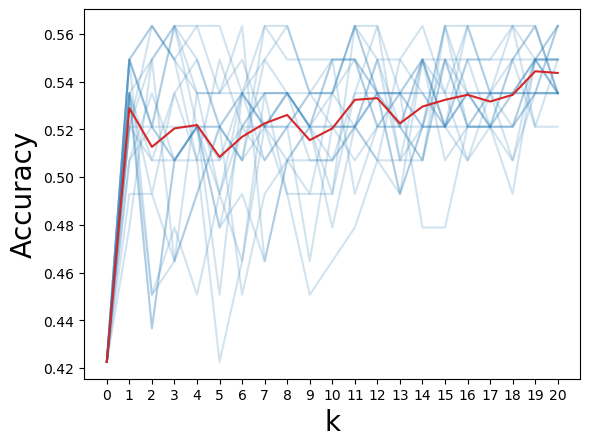}
            \caption{Opt-1.3B}\label{fig:removal-image3}
        \end{subfigure}
        \begin{subfigure}{0.24\linewidth}
            \centering
            \includegraphics[width=\textwidth]{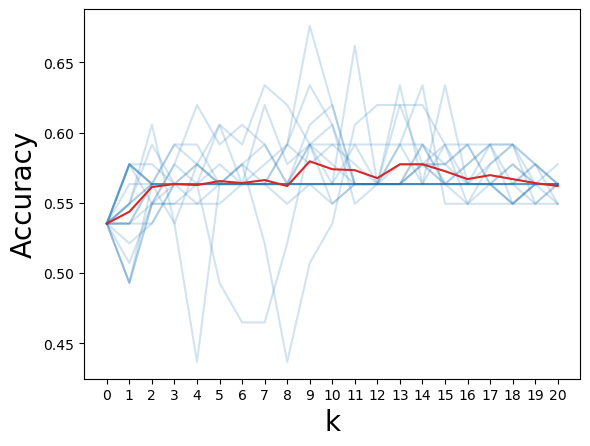}
            \caption{GPT-Neo-1.3B}\label{fig:removal-image4}
        \end{subfigure}%
        \hfill
        \begin{subfigure}{0.24\linewidth}
            \centering
            \includegraphics[width=\textwidth]{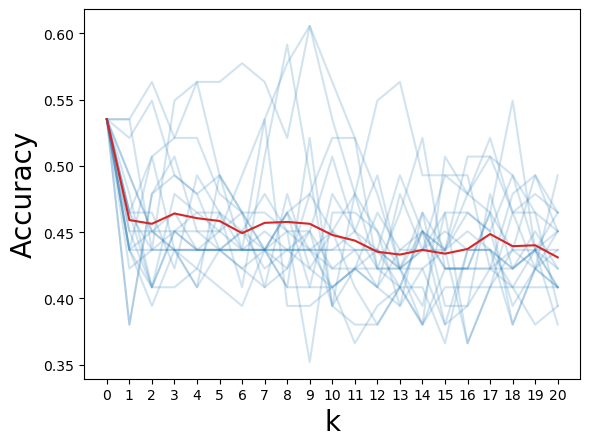}
            \caption{Pythia-1.4B}\label{fig:removal-image5}
        \end{subfigure}
        \hfill
            \begin{subfigure}{0.24\linewidth}
            \centering
            \includegraphics[width=\textwidth]{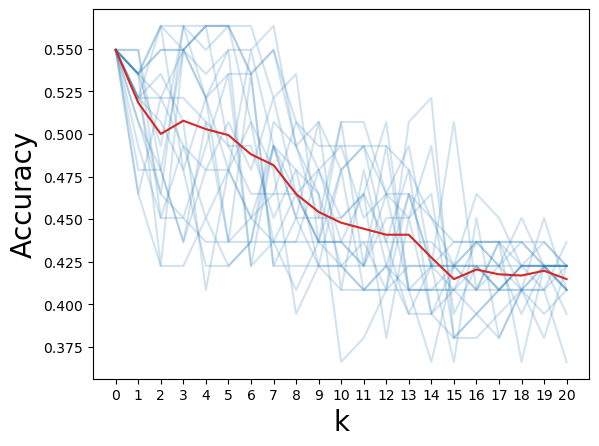}
            \caption{GPT-Neo-2.7B}\label{fig:removal-image6}
        \end{subfigure}
        \hfill
        \begin{subfigure}{0.24\linewidth}
            \centering
            \includegraphics[width=\textwidth]{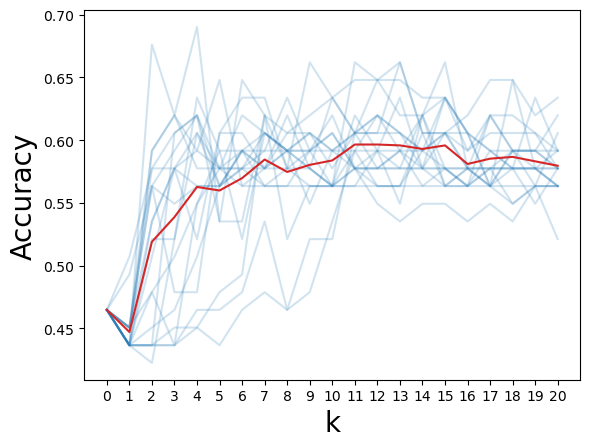}
            \caption{Llama2-7B}\label{fig:removal-image7}
        \end{subfigure}%
        \hfill
        \begin{subfigure}{0.24\linewidth}
            \centering
            \includegraphics[width=\textwidth]{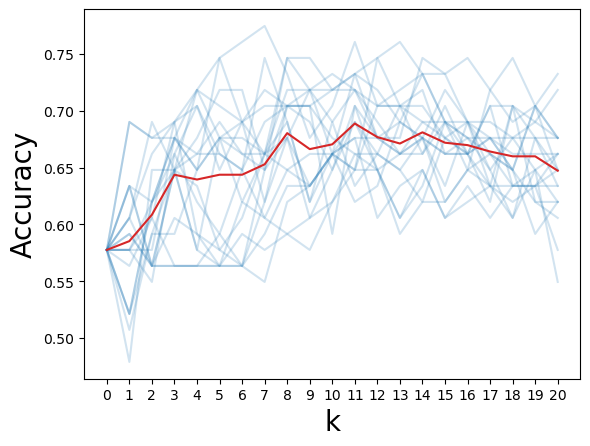}
            \caption{Llama2-13B}\label{fig:removal-image8}
        \end{subfigure}
        \hfill
            \begin{minipage}{.1cm}
            \vfill
            \end{minipage}
    \end{minipage}%
    \hfill
    \begin{minipage}[c]{\linewidth}
        \caption{\label{fig:}Performance of each model on WNLI dataset. In each plot, the red line indicates the averages of all permutations for one trial, overlaying blue lines for individual permutations.}
    \end{minipage}
\end{figure*}

\end{document}